\documentclass[10pt,twocolumn,letterpaper]{article}

\usepackage{cvpr}
\usepackage{times}
\usepackage{epsfig}
\usepackage{graphicx}
\usepackage{amsmath}
\usepackage{amssymb}
\usepackage{makecell}
\usepackage{amsmath}

\DeclareMathOperator*{\argmin}{arg\,min}

\usepackage{nopageno}

\usepackage[breaklinks=true,bookmarks=false]{hyperref}

\cvprfinalcopy % *** Uncomment this line for the final submission

\def\cvprPaperID{4637} % *** Enter the CVPR Paper ID here

\ifcvprfinal\fi
\begin{document}

%%%%%%%%% TITLE
\title{SDFDiff: Differentiable Rendering of Signed Distance Fields for 3D Shape Optimization}

\author{Yue Jiang, Dantong Ji, Zhizhong Han, Matthias Zwicker\\
University of Maryland, College Park\\
{\tt\small \{yuejiang, dji, h312h, zwicker\}@cs.umd.edu}
% For a paper whose authors are all at the same institution,
% omit the following lines up until the closing ``}''.
% Additional authors and addresses can be added with ``\and'',
% just like the second author.
% To save space, use either the email address or home page, not both
% \and
% Second Author\\
% % Institution2\\
% % First line of institution2 address\\
% {\tt\small secondauthor@i2.org}
}

\maketitle
%\thispagestyle{empty}

%%%%%%%%% ABSTRACT
\begin{abstract}
We propose SDFDiff, a novel approach for image-based shape optimization using differentiable rendering of 3D shapes represented by signed distance functions (SDFs). Compared to other representations, SDFs have the advantage that they can represent shapes with arbitrary topology, and that they guarantee watertight surfaces. We apply our approach to the problem of multi-view 3D reconstruction, where we achieve high reconstruction quality and can capture complex topology of 3D objects. In addition, we employ a multi-resolution strategy to obtain a robust optimization algorithm. We further demonstrate that our SDF-based differentiable renderer can be integrated with deep learning models, which opens up options for learning approaches on 3D objects without 3D supervision. In particular, we apply our method to single-view 3D reconstruction and achieve state-of-the-art results.
\end{abstract}

%%%%%%%%% BODY TEXT
\section{Introduction}
%Computer graphics studies the problem of generating images from descriptions of scenes. Rendering is a forward process used for generating 2D images from 3D objects. The rendering process itself is not differentiable because of discrete operations.  
The ``vision as inverse graphics'' or ``inverse rendering'' strategy has long been attractive as a conceptual framework to solve inverse problems such as recovering shape or appearance models from images. In this analysis-by-synthesis approach, the goal is to reproduce given input images by synthesizing them using an image formation model, possibly including shape, appearance, illumination, and camera parameters. Solving this problem implies finding suitable model parameters (shape, appearance, illumination, camera) that describe the underlying scene. While conceptually simple, this approach can be challenging to use in practice, because it requires a suitable parameterization of a powerful image formation model, and effective numerical techniques to solve the resulting optimization problem. Recently, automatic differentiation has attracted renewed attention to implement differentiable renderers or image formation models that can be used in gradient-based optimization techniques. In particular, it is attractive to combine differentiable rendering with neural networks to solve highly ill-posed inverse problems, such as single-view 3D reconstruction.

In this paper, we advocate using signed distance fields (SDFs) in a differentiable image formation model because they have several advantages over other geometry representations. In contrast to triangle meshes, the surface topology is not fixed in SDFs and can adapt to the actual scene topology during optimization. Point clouds can also represent arbitrary topologies but they do not provide continuous surface reconstructions. Instead, SDFs inherently represent continuous, watertight surfaces, which are required for downstream applications such as 3D printing and physics-based simulation. In addition, SDFs can easily be used in a multi-resolution framework, which is important to avoid undesired local minima during optimization.

The main contribution of our paper is SDFDiff, a differentiable renderer based on ray-casting SDFs. Our renderer is integrated with a deep learning framework such that it can be combined with neural networks to learn how to solve highly ill-posed inverse problems. Finally, we provide an effective multi-resolution strategy to improve the robustness of gradient-based optimization. We demonstrate the usefulness of our approach using several application studies, including multi-view 3D reconstruction and learning-based single-view 3D reconstruction. Our results demonstrate the advantages of SDFs over other surface representations.

In summary, we make the following contributions:
%contribute several innovative ideas in the area of 3D reconstruction:

\begin{itemize}
\item We introduce a differentiable renderer based on ray-casting SDFs, and we describe an implementation that is integrated into a standard deep learning framework. Advantages of using SDFs for differentiable rendering and shape optimization include that we can adapt the topology freely, and that the resulting shapes are guaranteed to consist of watertight surfaces. %Our SDF-based differentiable renderer can tranform between objects with significant topological differences.
\item We present results of a multi-view 3D reconstruction approach using shape optimization via differentiable rendering. Using a multi-resolution approach, our gradient-descent optimization reliably converges to high quality solutions. %, seamlessly integrating iterative optimization algorithms into standard 3D reconstruction methods. 
Our approach is able to reconstruct geometry with high level of detail and complex topology, even with few input views.
\item We leverage the SDF-based differentiable renderer to train deep neural networks to perform single-view 3D shape reconstruction without 3D supervision. We demonstrate the advantages of our approach by recovering accurate 3D shapes with arbitrary topology.

%algorithm can be applied to solve many vision as inverse graphics tasks. For example, we can generate realistic 3D objects without 3D supervision and perform unsupervised learning to construct 3D objects from sketches. 

\end{itemize}

\section{Related Work}

\paragraph*{Signed Distance Functions.}
A distance function is a level set representation \cite{Osher2004, Peng1999} that, at each point in 3D, stores the distance to the closest point on the 3D surface. Signed distance fields (SDFs)~\cite{Curless1996} store a signed  distance to distinguish between the inside and outside of objects. SDFs are often discretized using uniform voxel grids~\cite{Newcombe2011, Niessner2013}. Compared to meshes or parametric surfaces, the implicit surface representation~\cite{MeschederNetworks, chen2018implicit_decoder} of SDFs has the advantage that it can represent arbitrary topologies. In contrast to point clouds, SDFs always represent watertight surfaces. SDFs recently started attracting interest for shape analysis via deep learning. DeepSDF \cite{Park2019} was proposed to learn continuous SDF representation of a class of shapes. Similarly, deep level sets \cite{DBLP:journals/corr/abs-1901-06802} were introduced as an end-to-end trainable model that directly predicts implicit surfaces of arbitrary topology. However, these methods require 3D supervision during training, such as pairs of 3D coordinates and their corresponding SDF values \cite{Park2019}, or voxel occupancy maps~\cite{DBLP:journals/corr/abs-1901-06802}.

\paragraph*{Learning-based 3D Reconstruction.}
Reconstructing 3D models from 2D images is a classic problem in computer vision. Compared to traditional multi-view stereo and shading-based approaches \cite{seitz2016comparison}, learning-based 3D reconstruction can achieve impressive performance even with very few input views. Deep learning models for 3D shape understanding have been proposed for different kinds of 3D representations, including multiple views~\cite{Zhizhong2018VIP,Zhizhong2019seq}, point clouds~\cite{Yifan:DSS:2019, InsafutdinovD18}, triangle meshes~\cite{Liu:Paparazzi:2018, liu2019softras}, voxel grids~\cite{yan2016perspective, TulsianiZEM17}, and signed distance functions (SDFs)~\cite{Park2019}. However, most learning-based methods \cite{wang2018first, tulsiani2017learning, fan2017point, achlioptas2017representation, choy20163d, Jiang2018gal, kurenkov2018deformnet} require 3D supervision. Although some methods \cite{2018arXiv180910820C, 1b192a1631f4472b9980c2c84d0c5929} do not require supervised learning, they are often limited by specific settings, such as restricted lighting conditions or annotation of object orientation. In contrast, predicting 3D shapes with differentiable renderers \cite{Li2018, InsafutdinovD18, Jiang2018gal, navaneet2019capnet, liu2019dist, Niemeyer2020CVPR} has recently attracted increasing attention as it enables 3D reconstruction without 3D supervision, that is, by optimizing neural networks only using images as training data.

\paragraph*{Differentiable Rendering}
Voxel-based differentiable renderers~\cite{TulsianiZEM17, gadelha20163Dshape, henzler2019escaping, rezende2016unsupervised, yan2016perspective, gwak2017weakly, zhu2018von, wu2017marrnet, RenderNet2018} first drew attention performing volumetric ray marching, however, they are limited to low-resolution voxel grids. To render SDFs we use sphere tracing, also used in scene representation networks (SRNs)~\cite{Sitzmann:2019:SRN}. SRNs learn how to sphere trace novel views, but cannot produce full 3D shapes in a single step. They do not reconstruct a view independent shape representation and focus more on novel view synthesis rather than 3D watertight surface reconstruction, which is our goal. Starting with Loper and Black's OpenDR~\cite{Loper2014}, much recent work focused on mesh-based differentiable renderers \cite{1b192a1631f4472b9980c2c84d0c5929, palazzi2018end}. Kato et al.~\cite{kato2018} proposed a neural 3D mesh renderer with hand-designed gradients. 
%These two methods focus on the primary visibility gradients using triangle meshes as geometric representations. 
Paparazzi~\cite{Liu:Paparazzi:2018} employed analytically computed gradients to adjust the location of vertices. Similarly, SoftRas~\cite{liu2019softras} assigned each pixel to all faces of a mesh in a probabilistic rasterization framework. Although these methods enable to learn 3D mesh reconstruction without 3D supervision, they are restricted to a fixed, usually spherical mesh topology. Many of these mesh-based rendering approaches are differentiable with respect to geometry \cite{2018arXiv180910820C, kato2018, Loper2014, liu2018adversarial, delaunoy2011gradient, richardson2017learning, liu2019softras}, lighting models \cite{NIPS2019_9156}, textures \cite{liu2018adversarial, ramamoorthi2001signal, pmlr-v80-athalye18b}, or materials \cite{ramamoorthi2001signal, Meka2018lime, pmlr-v80-athalye18b}. Mesh-based differentiable rendering~\cite{peterson2019pix2vex,Li2018} has also been applied to real images, although current results are rather limited and applying differentiable rendering for real photographs remains an open challenge as it requires a comprehensive global illumination model.

Differentiable rendering has also been applied to Monte Carlo ray tracing \cite{Li2018} and point cloud rendering \cite{Jiang2018gal, navaneet2019capnet}. Insafutdinov et al.~\cite{InsafutdinovD18} proposed a point cloud-based differentiable renderer with visibility modeling by conducting orthogonal projection on voxelized 3D space holding the point cloud. Surface splatting~\cite{Yifan:DSS:2019} was employed to model the visibility in point cloud rendering. Although point clouds can be easily acquired using range sensing technology, such as Microsoft Kinect and LIDAR, they do not explicitly represent topology and require post-processing to produce watertight surfaces. Concurrent works \cite{liu2019dist, Niemeyer2020CVPR} proposed differentiable rendering based on SDFs and on occupancy networks, further improving the quality of 3D reconstruction.

\section{Overview}
\label{sec:overview}

% \begin{figure}[t]
% \includegraphics[width=\columnwidth]{3.png}
% \caption{Ray tracing process}
% \label{fig:}
% \end{figure}

%\subsection{Overview}

We propose a novel approach for image-based 3D shape optimization by leveraging SDFs as the geometric representation to perform differentiable rendering. Given a set of parameters $\Theta$ representing the geometry description, lighting model, camera position, etc, a renderer $R$ can be written as a forward operator that produces an image $I$ by computing $I = R(\Theta)$. In contrast, optimizing geometry and other scene parameters from images is a backward process. Given a desired target image $I$, our goal is to get the set of parameters $\Theta = R^{-1}(I)$ that produces the target image. The rendering process itself is not invertible. Hence, instead of solving the inverse rendering problem directly, we can formulate it as an energy minimization problem,
\begin{equation}
\label{eq:invrendering}
\Theta^* =  \argmin_{\Theta} \mathcal{L}_{\mathrm{img}}(R(\Theta),I) + \mathcal{L}_{\mathrm{reg}}(\Theta)
\end{equation}
where $\mathcal{L}_{\mathrm{img}}$ is a loss function measuring the distance between the target image and the rendered image from the 3D object. In practice, the loss is typically accumulated over multiple target images. Getting the desired parameters $\Theta^*$ is equivalent to minimizing the loss $\mathcal{L}$. While all rendering parameters including geometry, illumination, camera pose, and surface appearance could in theory be recovered from images this way, we focus on shape optimization in this paper and assume the other parameters are known.
To enable gradient-based optimization, a key issue is to obtain the gradient of $\mathcal{L}_{\mathrm{img}}(R(\Theta),I)$ with respect to the parameters $\Theta$. A differentiable renderer achieves this by producing not only images from a description of the scene, but also the derivatives of pixel values with respect to scene parameters. 

In this paper, we propose a novel differentiable renderer which uses signed distance functions (SDFs) and camera pose as inputs and renders an image. Our SDF-based differentiable renderer leverages the ray casting algorithm and uses automatic differentiation to compute the derivatives. 

%The signed distance function is represented using a 3D Cartesian grid. We use finite differencing  to compute normal vectors. Linear interpolation is used to continuously reconstruct the SDF from the discrete grid.

%By doing ray tracing, we can find the zero-crossing points and use the 8 vertices around it to interpolate the normal vectors corresponding to the zero-crossing points via finite differencing approach and do shading based on the normal vector. We would update the SDF values to minimize the error on the pixel level.

%We also present a multi-resolution optimizer for optimizing the output images. To get a good image, the SDF should have similar resolution as pixels. Hence, we start with a low-resolution image and a low-resolution SDF, i.e., start optimizing a coarse SDF in a low image resolution to get an overall shape, and then refine it to get better results.

\section{Differentiable SDF Rendering}
\label{sec:sdfraycasting}

We represent discrete SDFs by sampling SDF values on regular grids, and apply a standard ray casting algorithm based on sphere tracing~\cite{Hart1996} to find the intersection points between rays and the object surface. For this purpose we employ trilinear interpolation to reconstruct continuous SDFs that can be evaluated at any desired location. This allows us to continuously represent the object surface, which is given by the zero level set of the interpolated SDF.

A key observation is that the derivatives of a given pixel with respect to rendering parameters only depend on a local neighborhood of eight SDF samples that define the value of the trilinearly interpolated SDF at the surface intersection point. In other words, the sphere tracing process itself does not need to be differentiable. Instead, only the local computations involving the local set of eight SDF samples around the surface intersection need to be differentiable. Therefore, our approach proceeds in two stages: first, we apply sphere tracing to identify the eight samples nearest to the surface intersection. This step is not differentiable. Second, we locally compute the pixel color based on the local set of SDF samples. This step is implemented using an automatic differentiation framework to obtain the derivatives.

%A figure to illustrate the approach would be useful.

While differentiable ray marching of voxel grids has been used before~\cite{yan2016perspective, zhu2018von, RenderNet2018, simone}, these approaches are based on voxel opacities, given by binary variables or continuous occupancy probabilities. In these cases ray marching through the entire volume needs to be differentiated because all voxels along a ray may influence the corresponding pixel. 
%In addition, in these techniques the computation of surface normals and evaluation of shading models is not as straightforward as with SDFs.

\paragraph{Sphere Tracing.}

We perform ray casting via sphere tracing~\cite{Hart1996} in the first stage by starting from the ray origin, and evaluating the SDF using trilinear interpolation to find the minimum distance from the current point on the ray to the object. Then we move along the ray by that distance. Moving along the ray by the minimum distance to the object guarantees that we will never move across the boundary of the object, while allowing us to make a possibly large step towards the surface.  We repeat this process until we reach the surface of the object, that is, until the SDF value at our current position on the ray is small enough, or until we leave the bounding box of the object.  While the efficiency of sphere tracing can be improved by increasing the step size~\cite{liu2019dist}, we implemented sphere tracing directly in CUDA without support for automatic differentiation. Hence, the computation cost of this step is negligible in our approach.
%and only the last sphere tracing step is performed in a differentiable manner.

% MZ: I think the bounding box criterium makes more sense. Doesn't matter if it is implemented differently at this point.
%we stop the process after a certain number of iterations if we don't hit anything.

\paragraph{Differentiable Shading.}

In the second stage, we compute the pixel color as a function of the local SDF samples that define the SDF at the intersection point, as determined by the first stage.
%In practice, 
These computations are implemented in a framework that supports automatic differentiation, allowing us to easily obtain the derivatives of the pixel. 
%
%Trilinear interpolation is equivalent to spline interpolation using piecewise linear basis functions. Each piecewise linear basis is defined by a single coefficient, which corresponds to the SDF value stored at a grid vertex. Trilinear interpolation at any location corresponds to the sum of eight basis functions centered at the eight closest grid points.
%
For each pixel, the input
%to this stage 
consists of the light and camera parameters, and the eight SDF samples closest to the ray-surface intersection point. 
%In our current implementation 
The computations include: getting the intersection point and the surface normal at the intersection point as a function of the trilinear basis coefficients (i.e., the eight SDF samples), and evaluating a shading model.

To take into account the dependence of the pixel value on the ray-surface intersection point, we express the intersection point as a function of the eight local SDF samples. Let us denote the local SDF values  by $d_0,\dots,d_7$, the current position on the ray (obtained from the ray casting stage) by $s \in \mathbb{R}^3$, and the unit ray direction by $v \in \mathbb{R}^3$. To express the intersection point as a function of $d_0,\dots,d_7$, we use the same approximation as in the ray casting stage, that is, the approximate intersection is $p(d_0,\dots,d_7) = s+\mathrm{trilinear}(d_0,\dots,d_7;s)v$, where $\mathrm{trilinear}(d_0,\dots,d_7;s)$ is the trilinear interpolation of the SDF at location $s$ and considered as a function of $d_0,\dots,d_7$. This approximation is conservative in the sense that it is accurate only if the SDF represents a plane that is perpendicular to the ray direction $v$. Otherwise, $p(d_0,\dots,d_7)$ is guaranteed not to cross the true intersection along the ray.

As an alternative to our conservative approximation, one could express the intersection point exactly as the solution of the intersection of the ray $s+tv$ and the local trilinear interpolation of the SDF. That is, we could express the solution of $\mathrm{trilinear}(d_0,\dots,d_7;s+tv)=0$ with respect to $t \in \mathbb{R}$ as a function of $d_0,\dots d_7$. However, this involves finding roots of a cubic polynomial, and we found that our much simpler approach works more robustly in practice.

To evaluate a shading model, we need the surface normal at the intersection point $p(d_0,\dots,d_7)$. Considering that the surface normal corresponds to the gradient of the SDF, we first compute gradients at the grid vertices using central finite differencing, and then trilinearly interpolate them at the intersection point $p(d_0,\dots,d_7; s)$. In summary, this leads to an expression of the normal at the intersection point as a function of SDF coefficients within an $4\times 4\times 4$ neighborhood around the intersection (because of central finite differencing). Surface normals are normalized after trilinear interpolation. Finally, in our current implementation we evaluate a simple diffuse shading model.

\paragraph{Implementation.}

We implemented SDF ray casting using CUDA to leverage the computational power of GPUs. Differentiable shading is implemented with the Pytorch library, which supports automatic differentiation and allows seamless integration of the renderer with 
%deep learning and 
neural network training. Pytorch also leverages the GPU, and our implementation directly accesses the output of the ray casting stage that is stored in GPU memory, avoiding any unnecessary memory transfers. Our code is available at https://github.com/YueJiang-nj/CVPR2020-SDFDiff.

\section{Multi-view 3D Reconstruction}

\begin{figure}[t]
\includegraphics[width=\columnwidth]{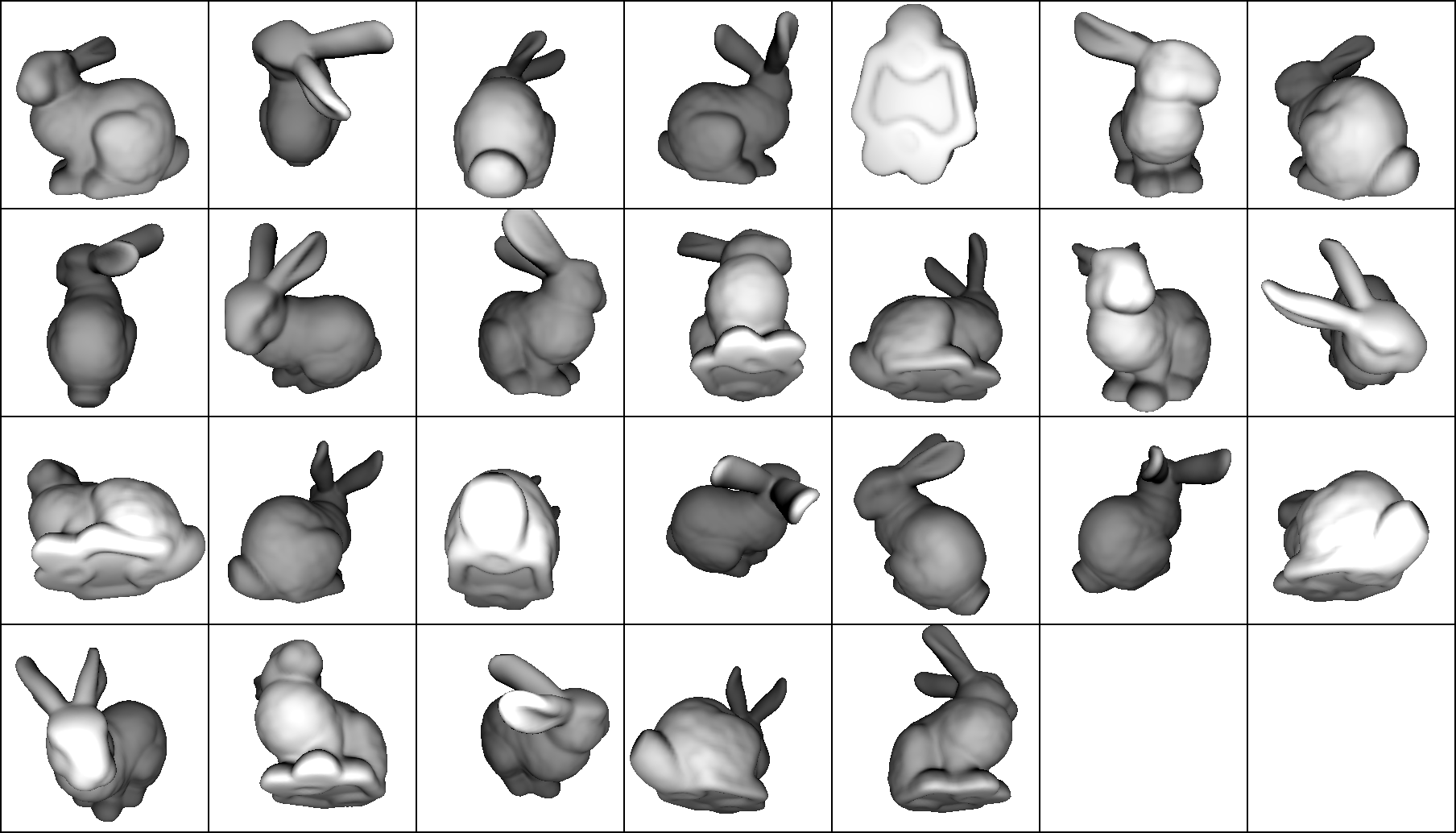}
\caption{We use 26 input views in our multi-view reconstruction experiments as shown here for the bunny.}
\label{fig:input}
\end{figure}

In this section we describe how to perform multi-view 3D reconstruction using our differentiable renderer. This is a proof of concept, where we assume known camera poses, illumination, and surface appearance, and we only optimize over the 3D shape represented by the SDF. %We first introduce the energy function that we minimize in Section~\ref{sec:energyfunction}, and then present a multi-scale optimization algorithm in Section~\ref{sec:optimization}.
%
%We also present a multi-resolution optimizer for optimizing the output images. To get a good image, the SDF should have similar resolution as pixels. Hence, we start with a low-resolution image and a low-resolution SDF, i.e., start optimizing a coarse SDF in a low image resolution to get an overall shape, and then refine it to get better results.
%
%\subsection{Input}
%
Our inputs are synthetically rendered images from a fixed set of camera poses. We set the camera poses to point from the center of each face and edge, and from each vertex of the  object bounding box towards its center, where the bounding box is a cube. Since the cube has 6 faces, 8 vertices, and 12 edges, we obtain 26 camera poses in total. \autoref{fig:input} shows the input images we used to reconstruct the bunny in our experiments. In addition, we initialize the SDF to a sphere.

\subsection{Energy Function}
\label{sec:energyfunction}

For simplicity we choose the $L_2$ distance between the rendered and the target images as our image-based loss, that is $\mathcal{L}_{\mathrm{img}}(R(\Theta),I)=||R(\Theta)-I||^2$. The loss is summed over all target views. In this proof of concept scenario, the optimization parameters $\Theta$ include only the SDF values, as we assume the other rendering parameters are known. Minimizing the image-based loss by optimizing SDF values requires differentiable rendering, where we compute the gradient of the image loss w.r.t. the SDF values as in Section~\ref{sec:sdfraycasting}.

In addition, we impose a regularization loss that ensures that the SDF values $\Theta$ represent a valid signed distance function, that is, its gradient should have unit magnitude. Writing the SDF represented by $\Theta$ as a function $f(x;\Theta)$, where $x$ is a point in 3D, the regularization loss is
\begin{align}
    \mathcal{L}_{\mathrm{reg}} = \int ||1-||\nabla f(x;\Theta)||^2||^2 dx
\end{align}
In practice, we obtain the gradients via finite differencing and we compute a discrete sum over the SDF grid vertices.

\subsection{Iterative Optimization}
\label{sec:optimization}

We apply gradient descent optimization using ADAM~\cite{kingma2014adam} to iteratively optimize our SDF to match the target images. Compared to straightforward gradient descent, ADAM is more robust and faster in convergence. In addition, we accelerate convergence by greedily selecting a single view in each gradient descent step to compute the gradient, similar to active mini batch sampling. The intuition is that some parts of the 3D model may have more complex structures so it is more difficult to optimize SDF values using some views than others. Different views may incur image losses of varying magnitude, and we should focus on the views with large losses to make sure all parts of the object can be well-reconstructed. Our approach first calculates the average loss for all the camera views from the result of the previous iteration. If a loss for a view is greater than the average loss, then during the current iteration, we update the SDF until the loss for this view is less than the average (with max. 20 updates). For the other views, we update the SDF five times. If one update increases the loss, then we switch to the next view directly. We stop our optimization process when the loss is smaller than a given tolerance or the step length is too small.

Reconstructing high-resolution 3D objects is challenging because gradient descent takes many iterations to eliminate low frequency errors. Therefore, we apply a coarse-to-fine multi-resolution approach. We start by initializing the SDF grid at a resolution of $8^3$ to the SDF of a sphere. We then iterate between performing gradient descent optimization as described above, and increasing the grid resolution. We increase the resolution simply by performing trilinear interpolation and stop at a resolution of $64^3$.

To further improve the efficiency of the multiresolution scheme, we choose an appropriate image resolution for rendering corresponding to the SDF resolution at each resolution level. We determine the appropriate resolution by ensuring that a sphere with a radius equivalent to the grid spacing, and placed at the corner of the bounding box of the SDF furthest from the camera, has a projected footprint of at most the size of a $2\times2$ pixel block.

\begin{figure}[t]
  \centering
  \resizebox{\columnwidth}{!}{
 \begin{tabular}{ | c | c | c | c | c | c | c | c | c | c | c | c | c | c | c | c |}
    \hline
    % Layout Pattern &   Time Comparison \\ \hline
    GT & 
    \begin{minipage}{0.2\textwidth}
      \includegraphics[width=40mm]{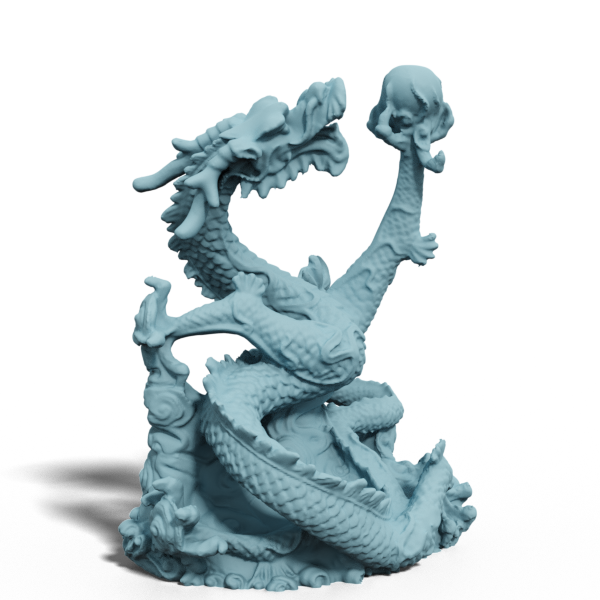}
      \vspace{0.01mm}
    \end{minipage}
    &
\begin{minipage}{0.2\textwidth}
      \includegraphics[width=40mm]{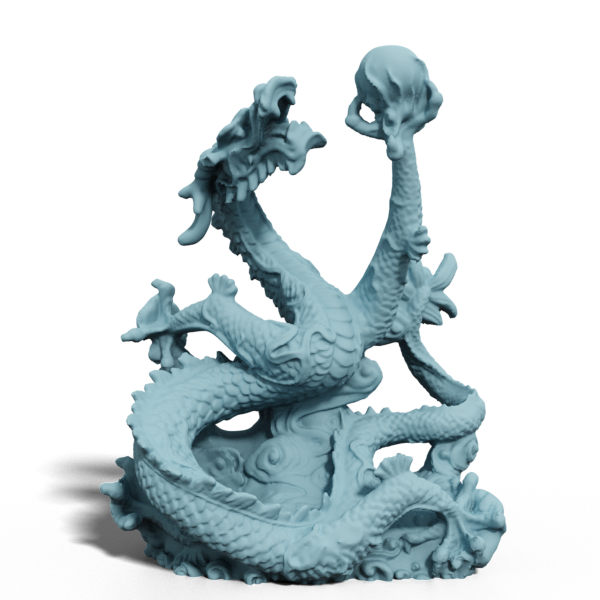}
      \vspace{0.01mm}
    \end{minipage}
     &
\begin{minipage}{0.2\textwidth}
      \includegraphics[width=40mm]{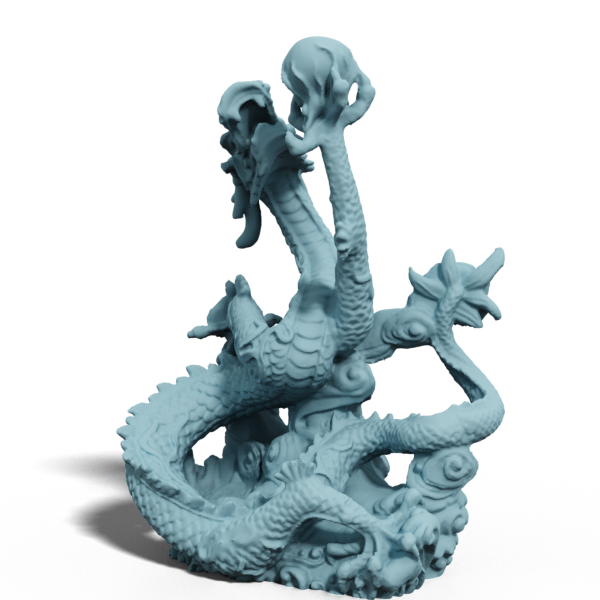}
       \vspace{0.01mm}
    \end{minipage}
     &
\begin{minipage}{0.2\textwidth}
      \includegraphics[width=40mm]{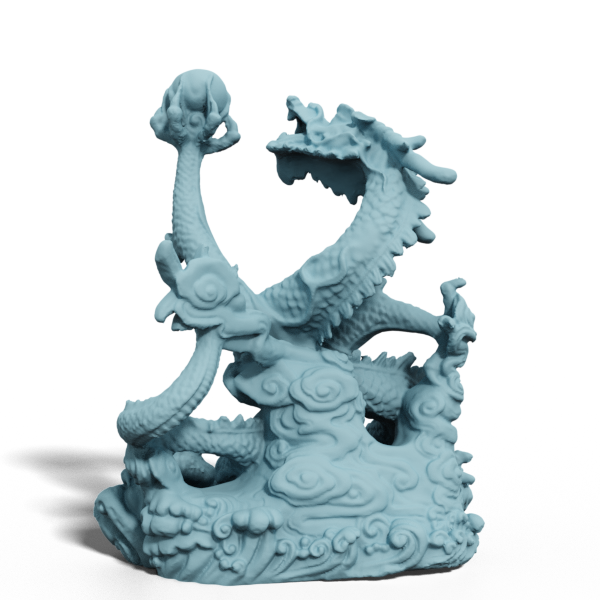}
      \vspace{0.01mm}
    \end{minipage}
  
    \\ \hline
 
   {\bf Ours} & 
   \begin{minipage}{0.2\textwidth}
      \includegraphics[width=40mm]{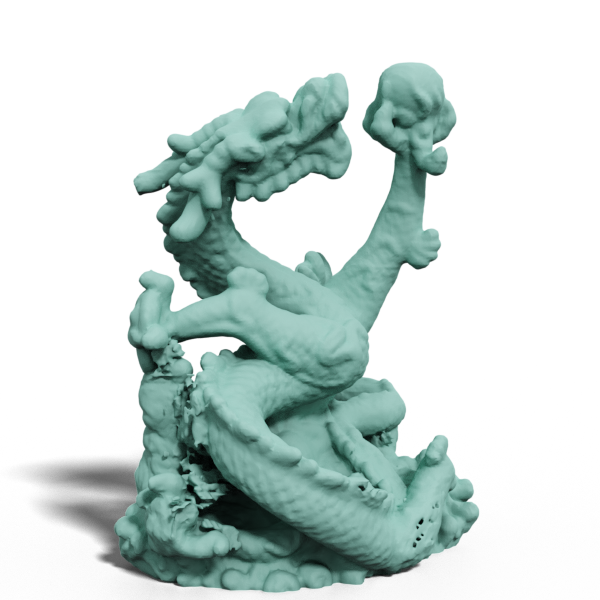}
      \vspace{0.01mm}
    \end{minipage}
    &
\begin{minipage}{0.2\textwidth}
      \includegraphics[width=40mm]{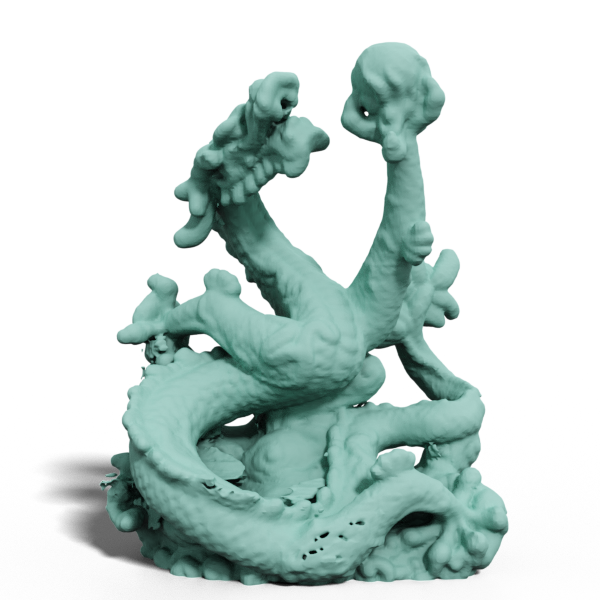}
      \vspace{0.01mm}
    \end{minipage}
     &
\begin{minipage}{0.2\textwidth}
      \includegraphics[width=40mm]{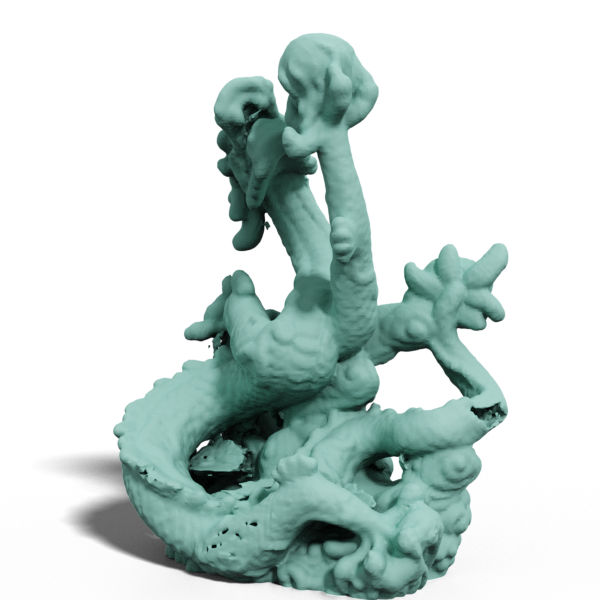}
      \vspace{0.01mm}
    \end{minipage}
     &
\begin{minipage}{0.2\textwidth}
      \includegraphics[width=40mm]{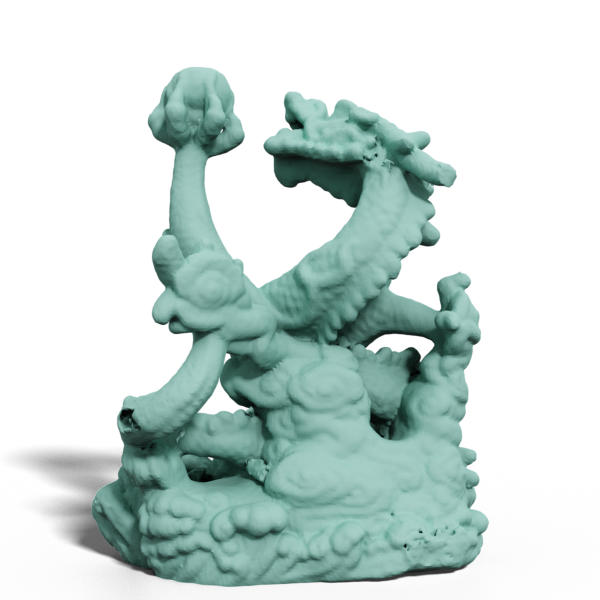}
      \vspace{0.01mm}
    \end{minipage}
    
      \\ \hline
    
     Error & 
    \begin{minipage}{0.2\textwidth}
      \includegraphics[width=40mm]{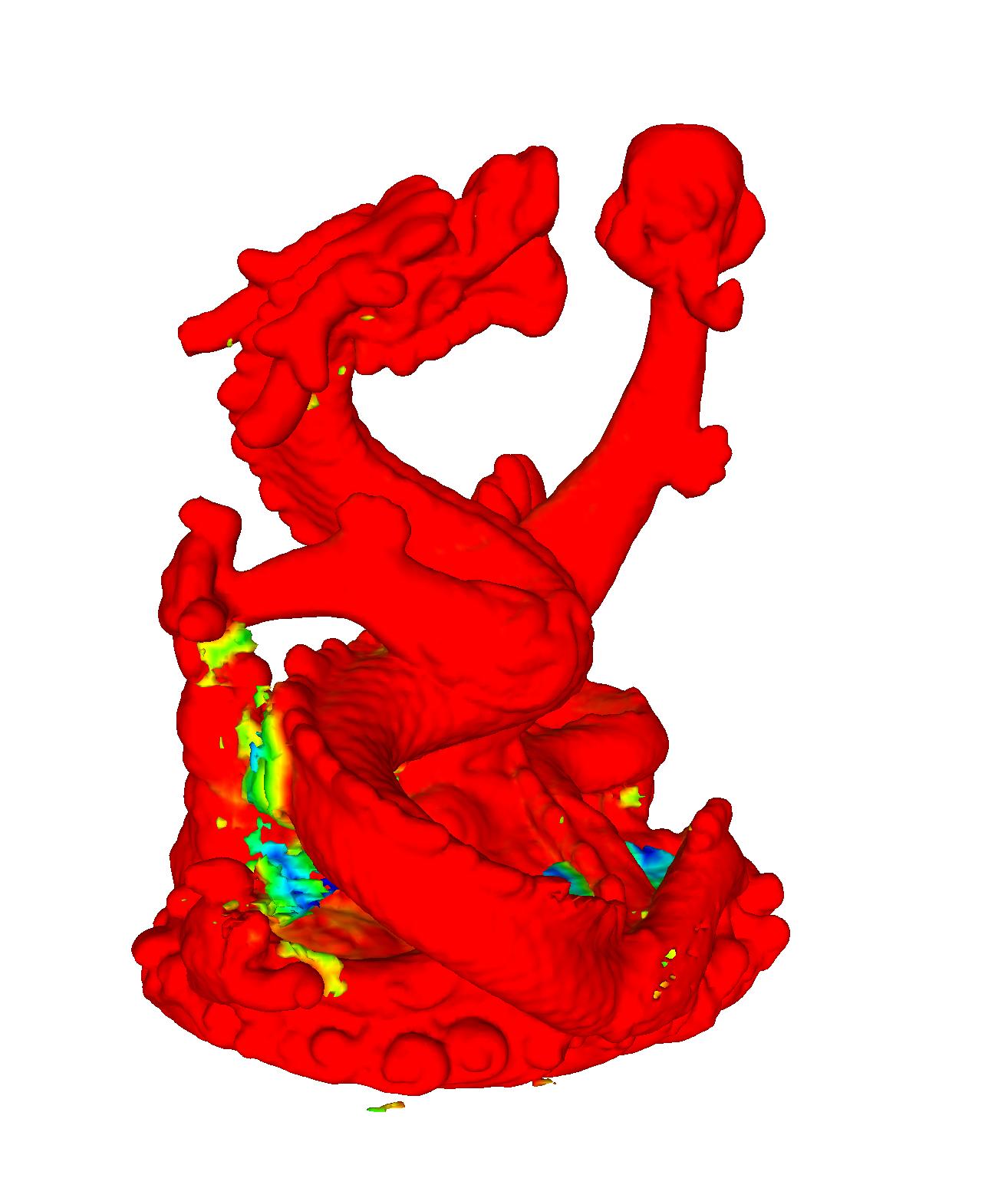}
    \end{minipage}
    &
\begin{minipage}{0.2\textwidth}
      \includegraphics[width=40mm]{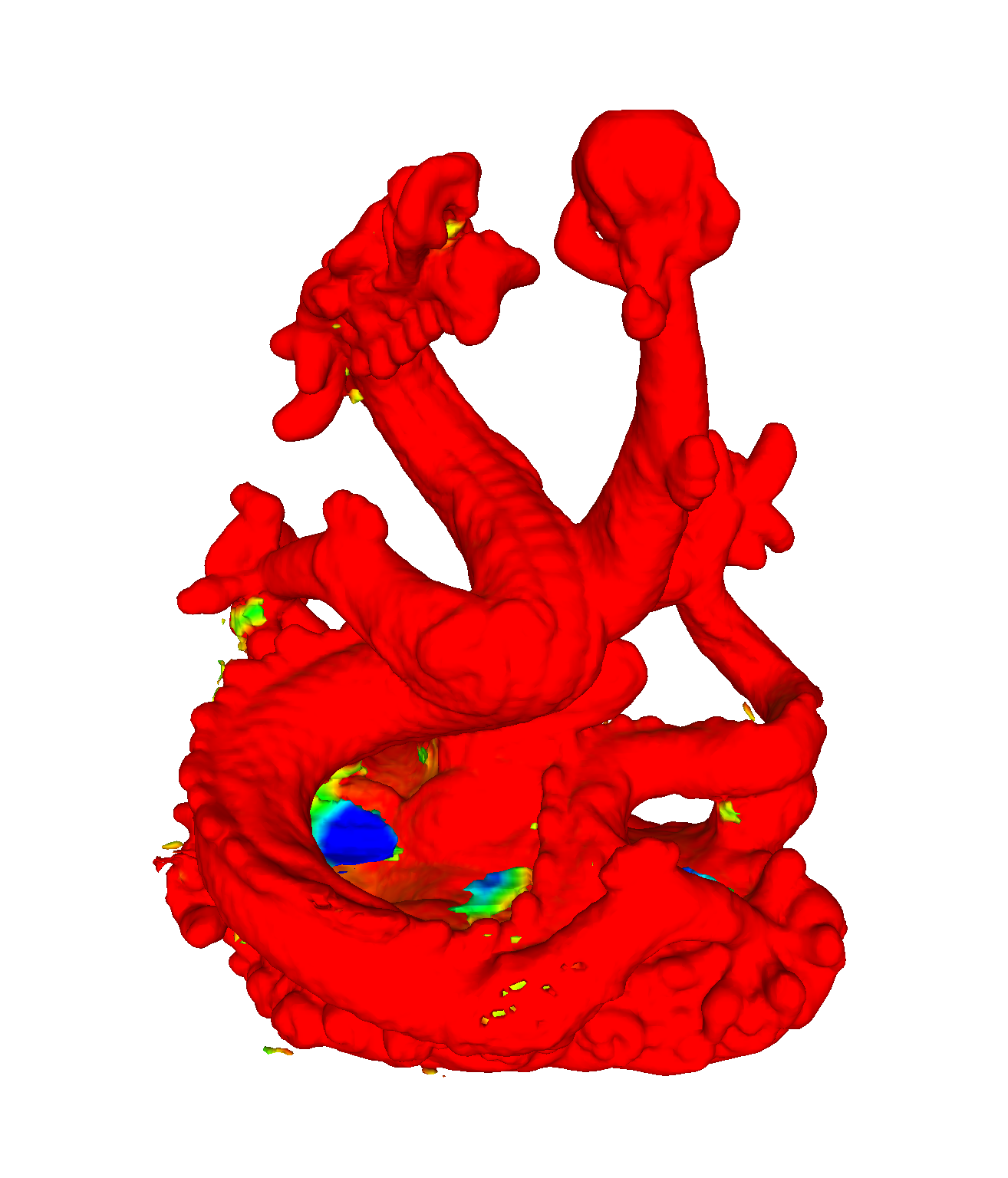}
    \end{minipage}
     &
\begin{minipage}{0.2\textwidth}
      \includegraphics[width=40mm]{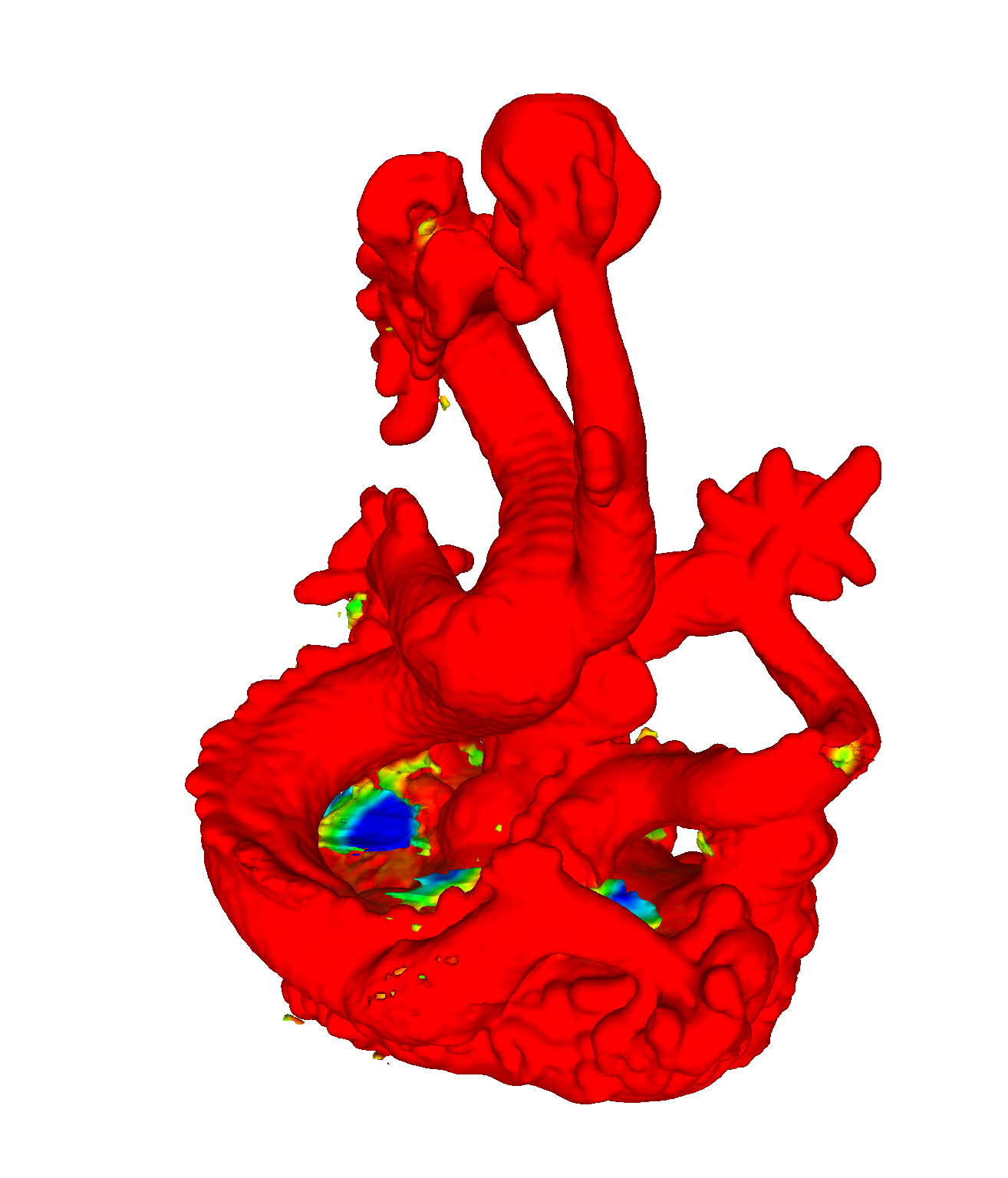}
    \end{minipage}
     &
\begin{minipage}{0.2\textwidth}
      \includegraphics[width=40mm]{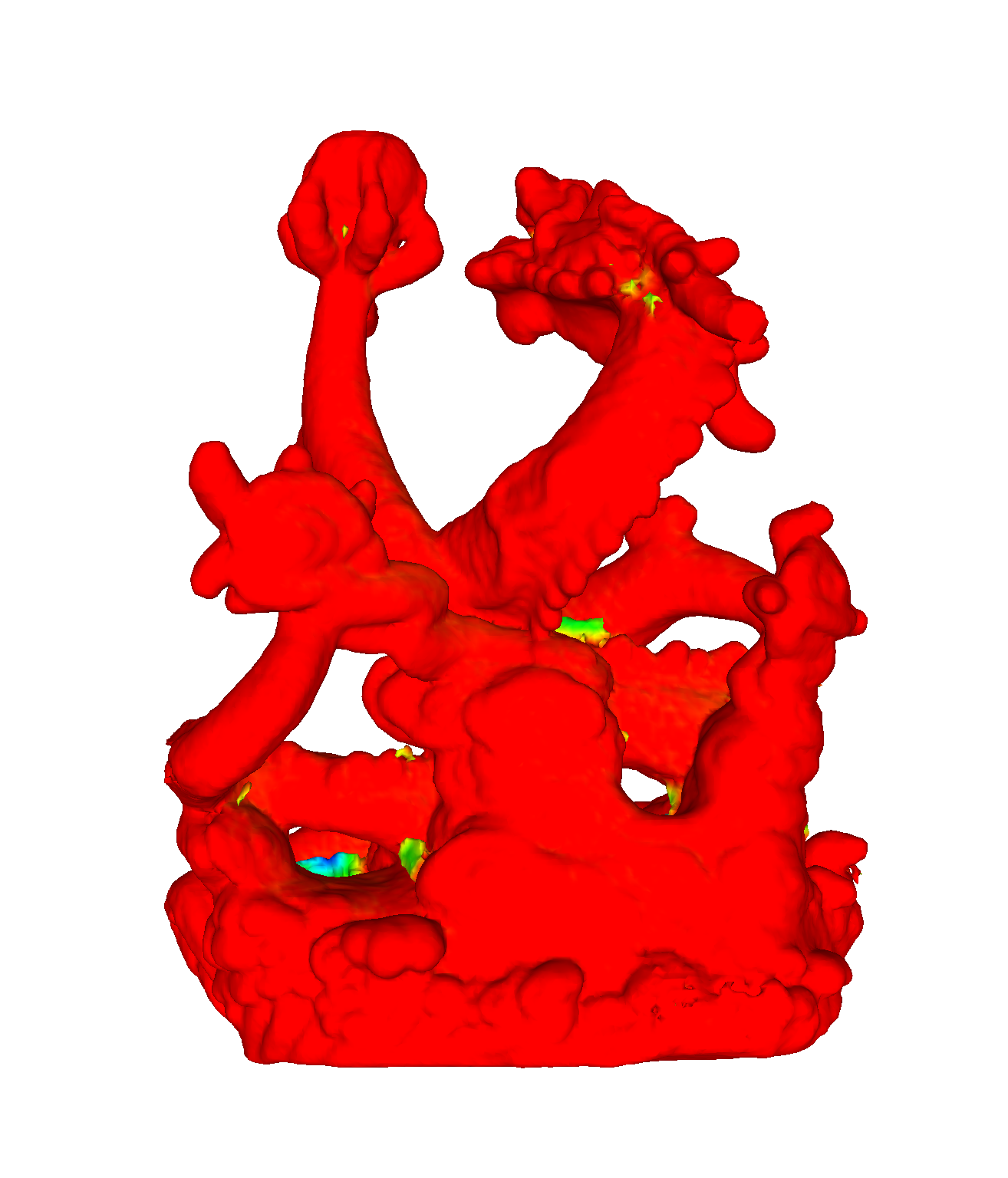}
    \end{minipage}
    
       \\ \hline

   \makecell{DSS\\ \cite{Yifan:DSS:2019}} & 
    \begin{minipage}{0.2\textwidth}
      \includegraphics[width=40mm]{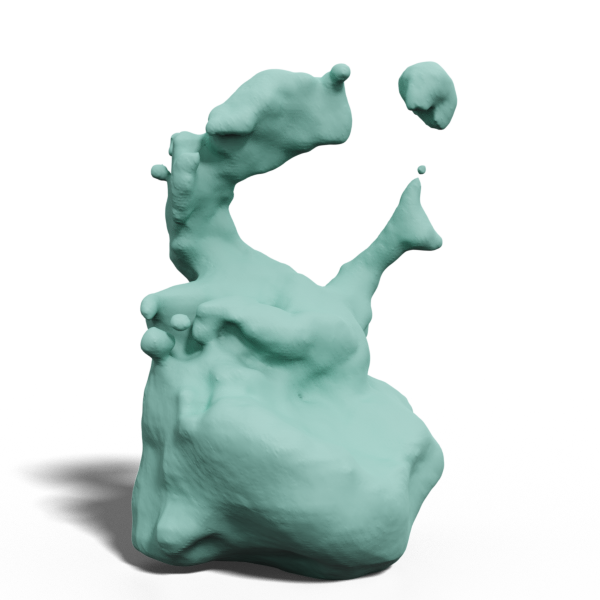}
    \end{minipage}
    &
\begin{minipage}{0.2\textwidth}
      \includegraphics[width=40mm]{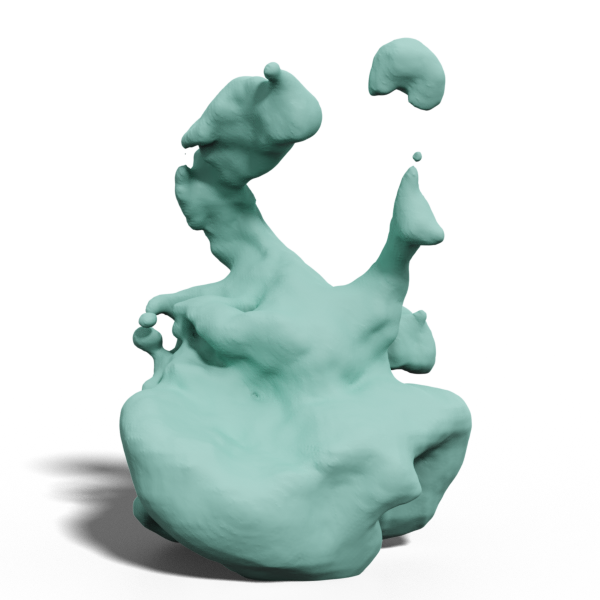}
    \end{minipage}
     &
\begin{minipage}{0.2\textwidth}
      \includegraphics[width=40mm]{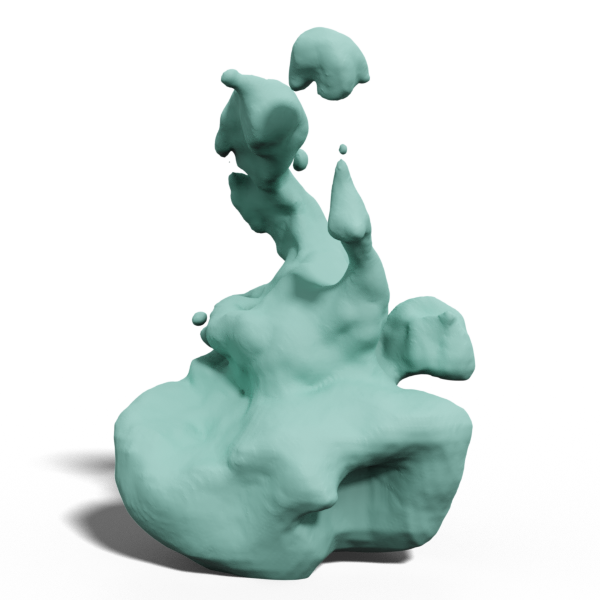}
    \end{minipage}
     &
\begin{minipage}{0.2\textwidth}
      \includegraphics[width=40mm]{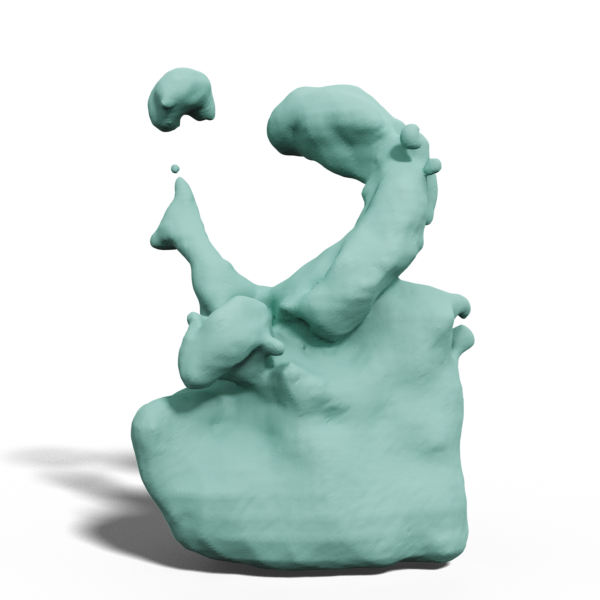}
    \end{minipage}
    
    \\ \hline
    
      Error & 
    \begin{minipage}{0.2\textwidth}
      \includegraphics[width=40mm]{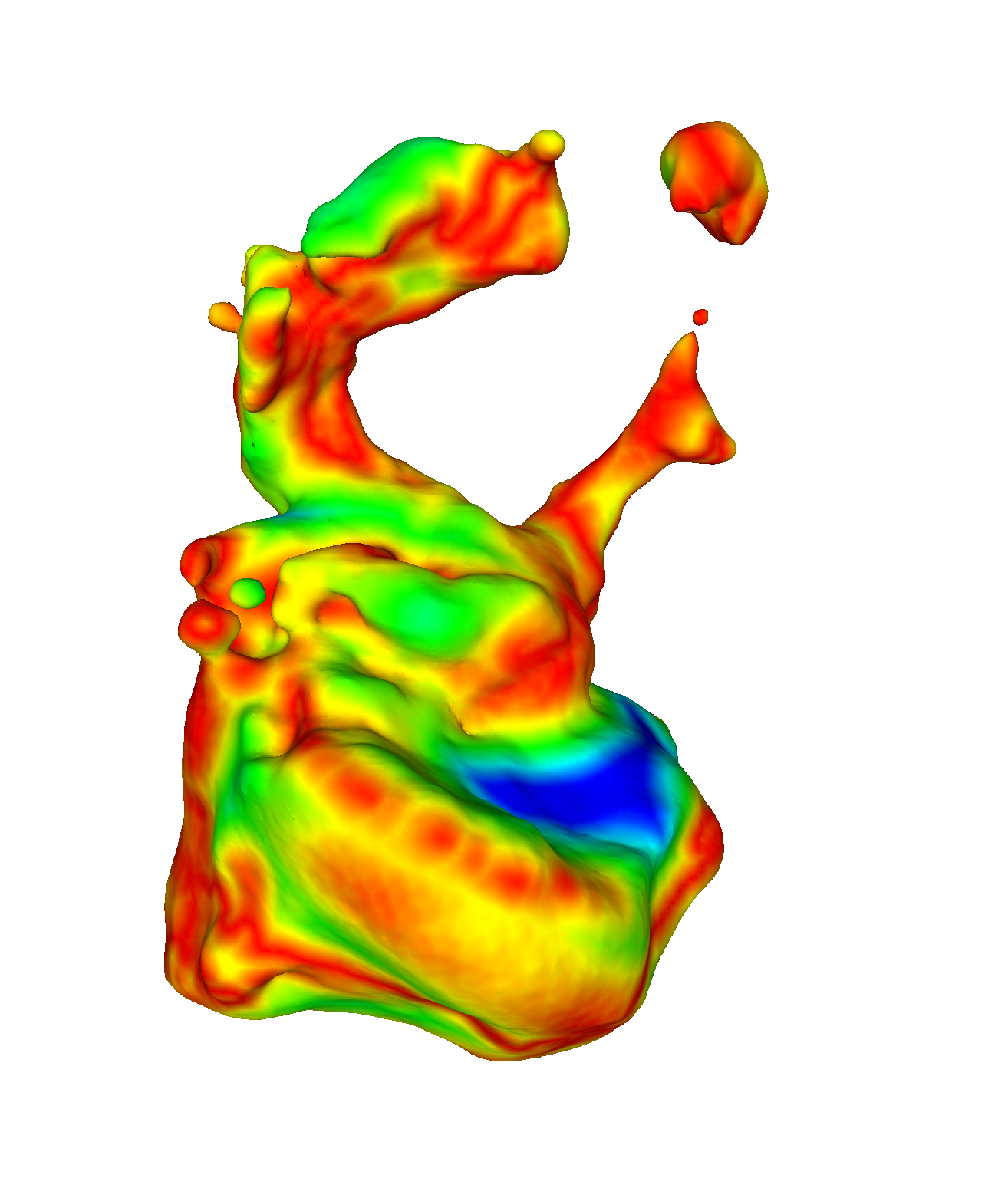}
    \end{minipage}
    &
\begin{minipage}{0.2\textwidth}
      \includegraphics[width=40mm]{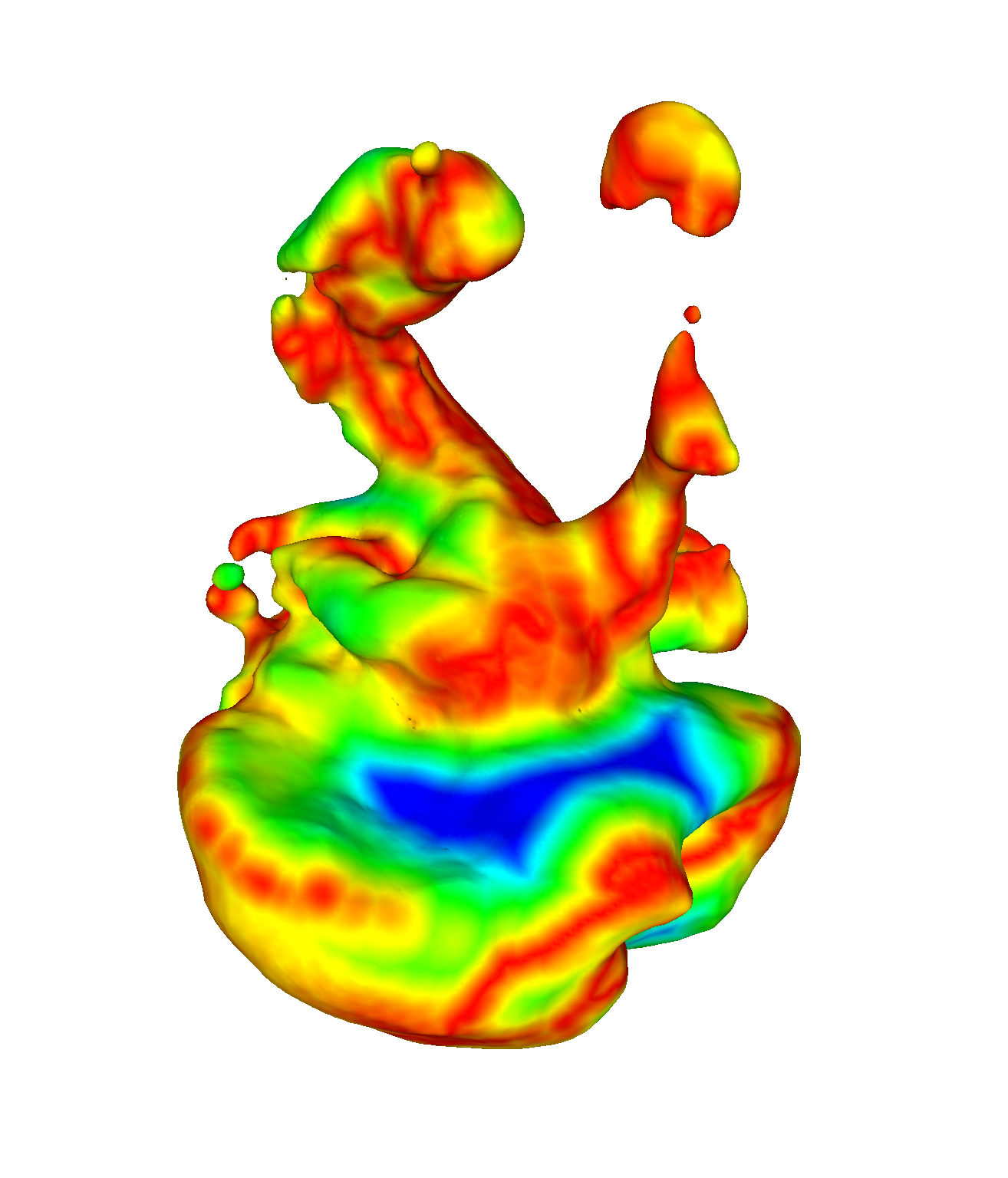}
    \end{minipage}
     &
\begin{minipage}{0.2\textwidth}
      \includegraphics[width=40mm]{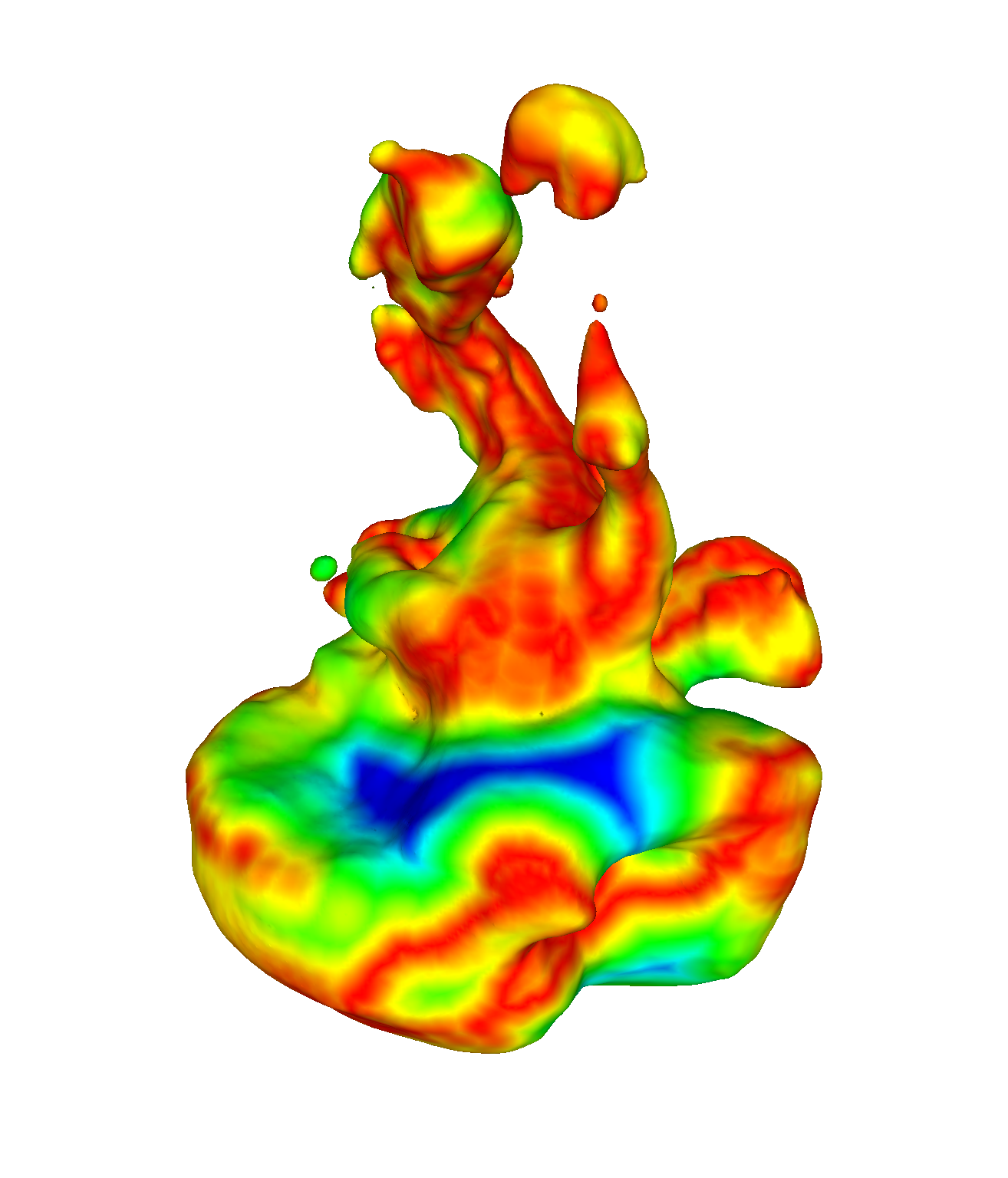}
    \end{minipage}
     &
\begin{minipage}{0.2\textwidth}
      \includegraphics[width=40mm]{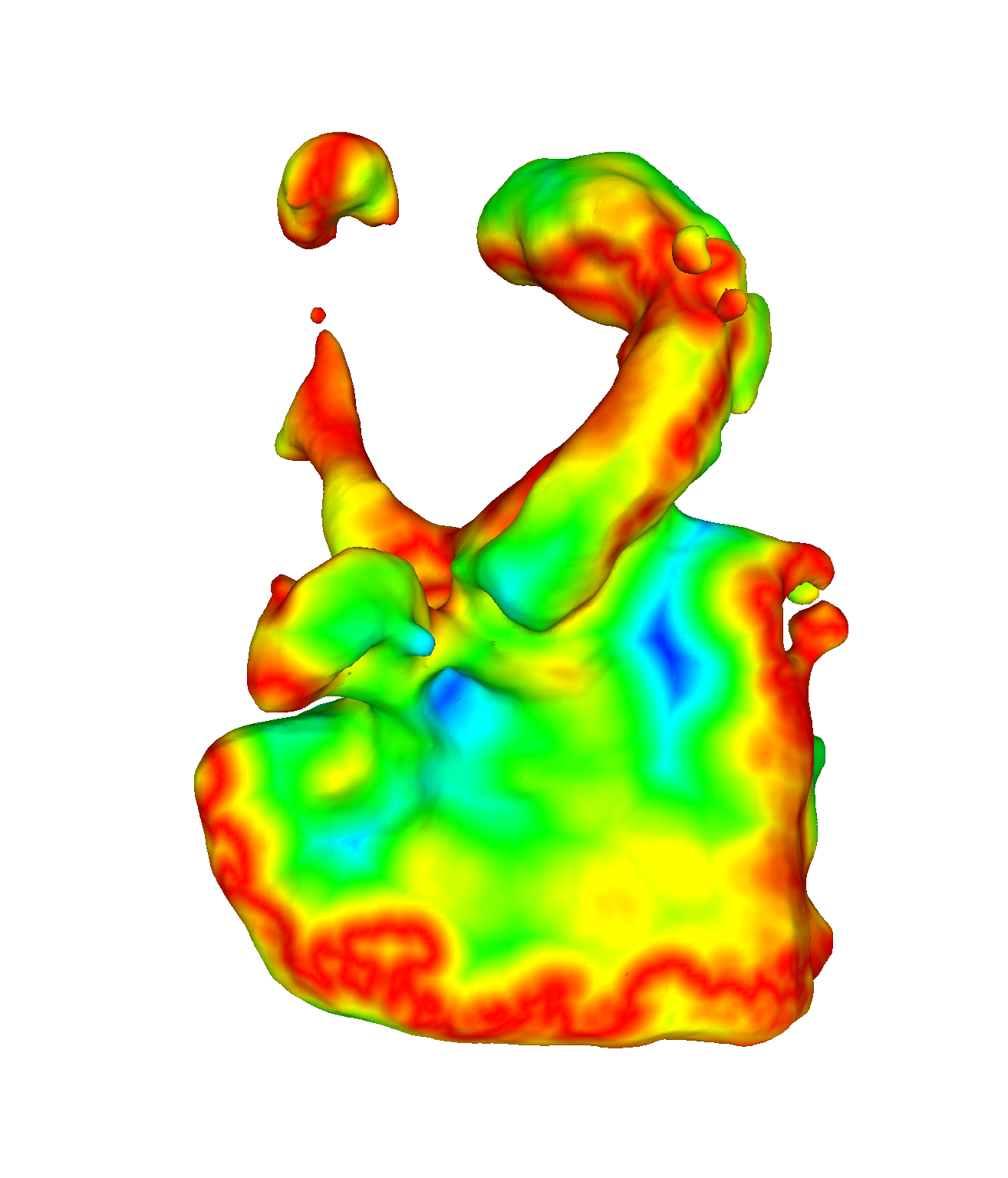}
    \end{minipage}
    
    \\ \hline

    \makecell{SMVS \\ \cite{langguth-2016-smvs, Semerjian2014}} & 
    \begin{minipage}{0.2\textwidth}
      \includegraphics[width=40mm]{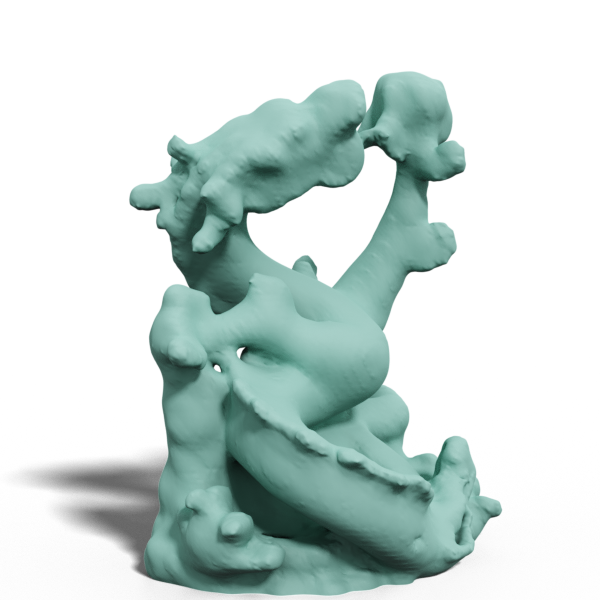}
      \vspace{0.01mm}
    \end{minipage}
    &
\begin{minipage}{0.2\textwidth}
      \includegraphics[width=40mm]{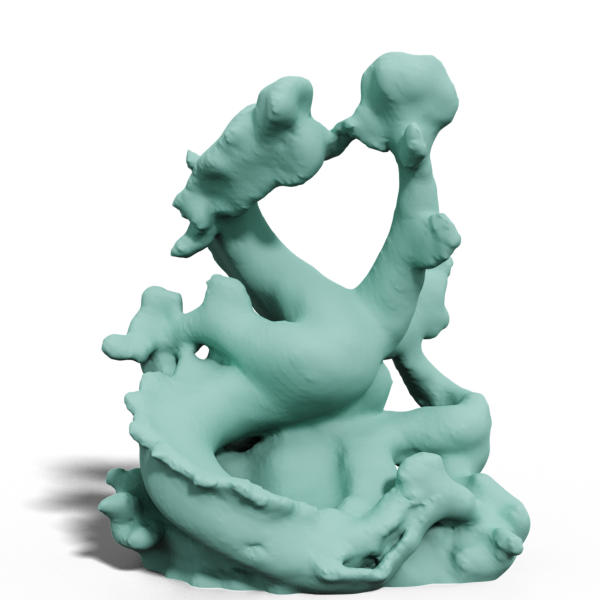}
      \vspace{0.01mm}
    \end{minipage}
     &
\begin{minipage}{0.2\textwidth}
      \includegraphics[width=40mm]{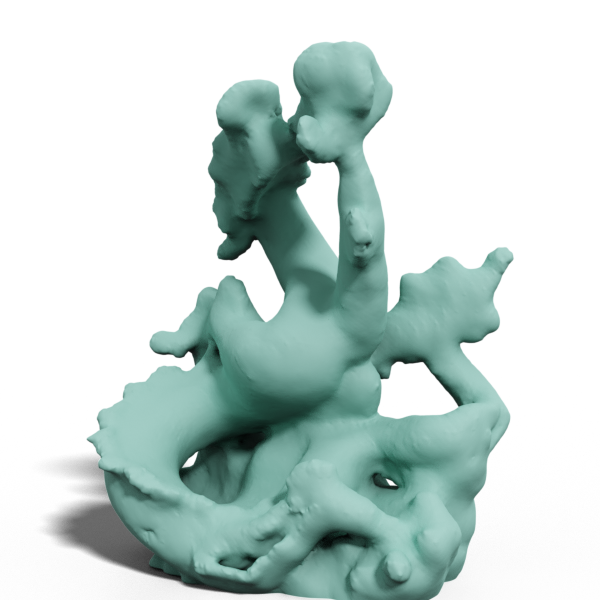}
      \vspace{0.01mm}
    \end{minipage}
     &
\begin{minipage}{0.2\textwidth}
      \includegraphics[width=40mm]{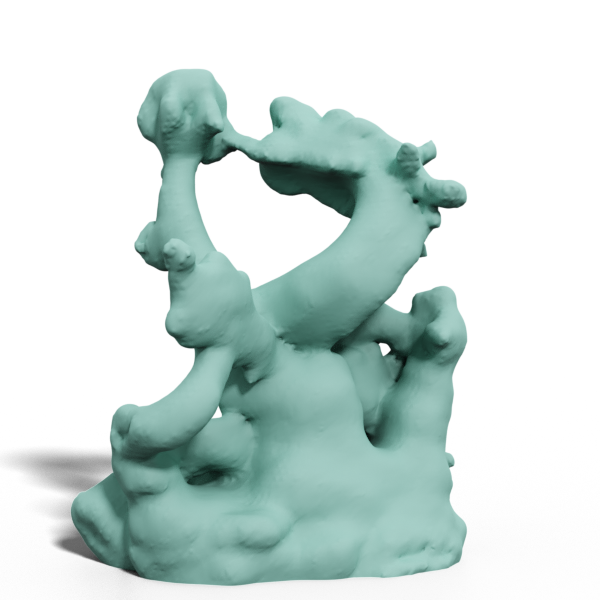}
      \vspace{0.01mm}
    \end{minipage}
    
  \\ \hline
    
      Error & 
    \begin{minipage}{0.2\textwidth}
      \includegraphics[width=40mm]{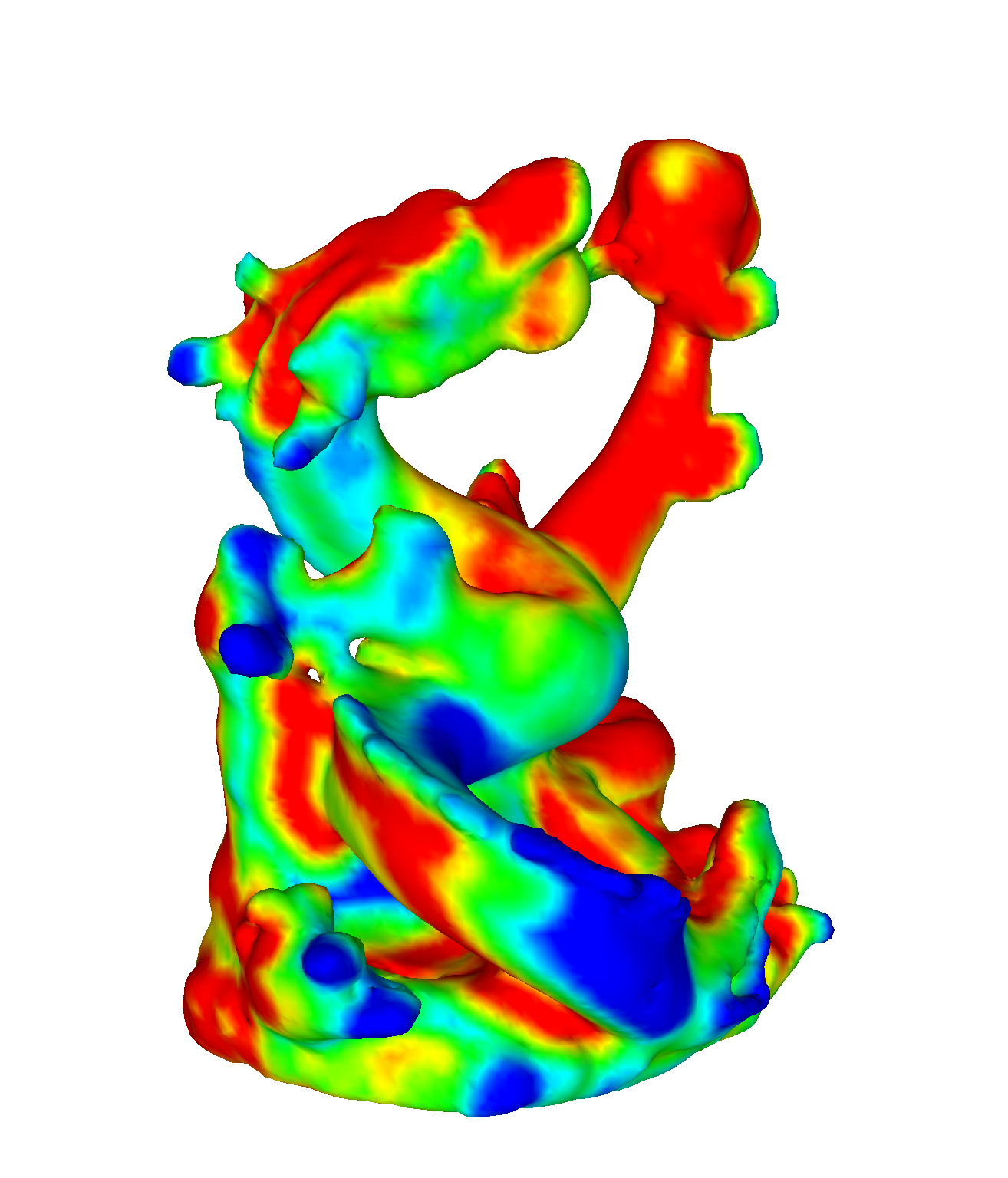}
    \end{minipage}
    &
\begin{minipage}{0.2\textwidth}
      \includegraphics[width=40mm]{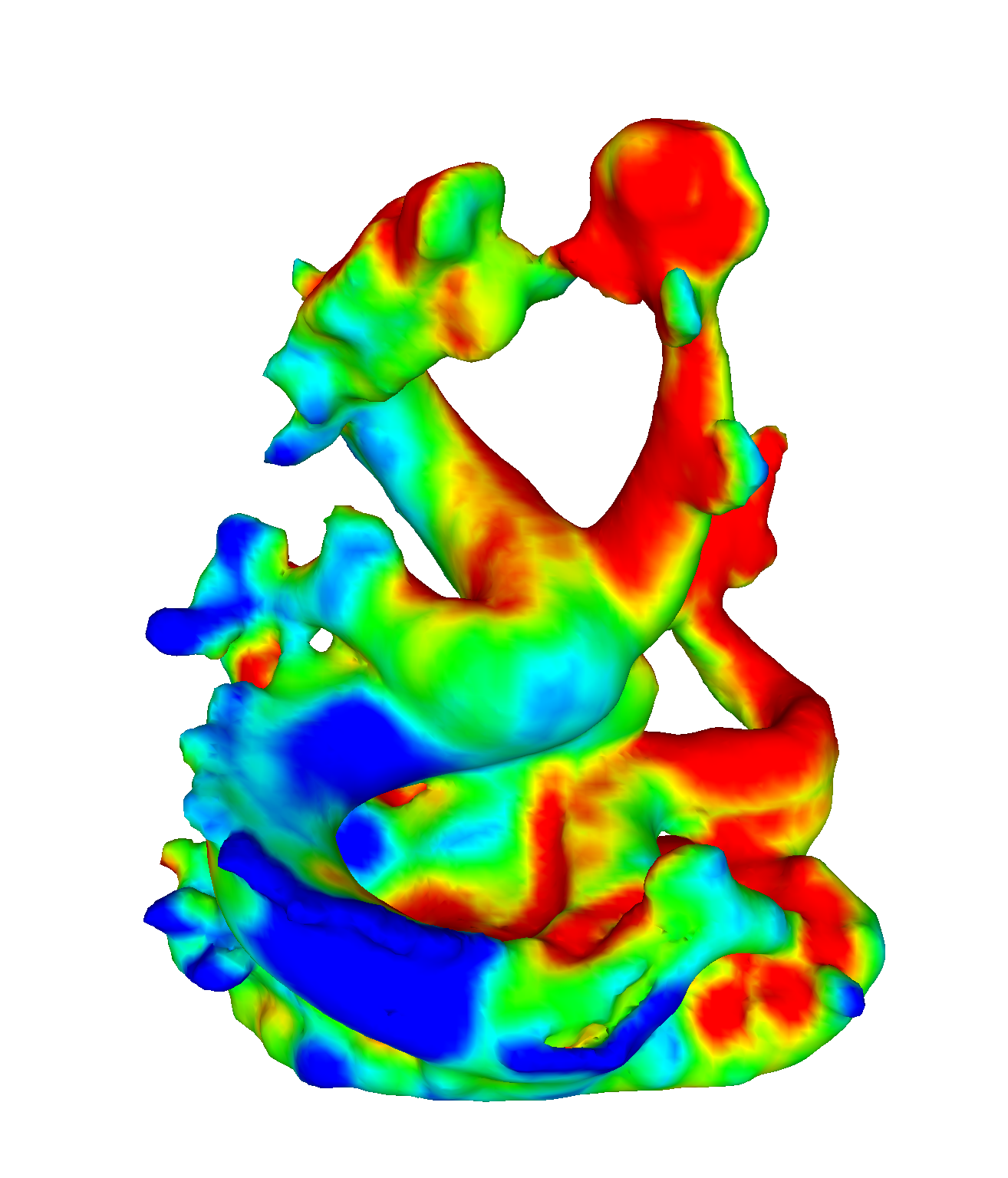}
    \end{minipage}
     &
\begin{minipage}{0.2\textwidth}
      \includegraphics[width=40mm]{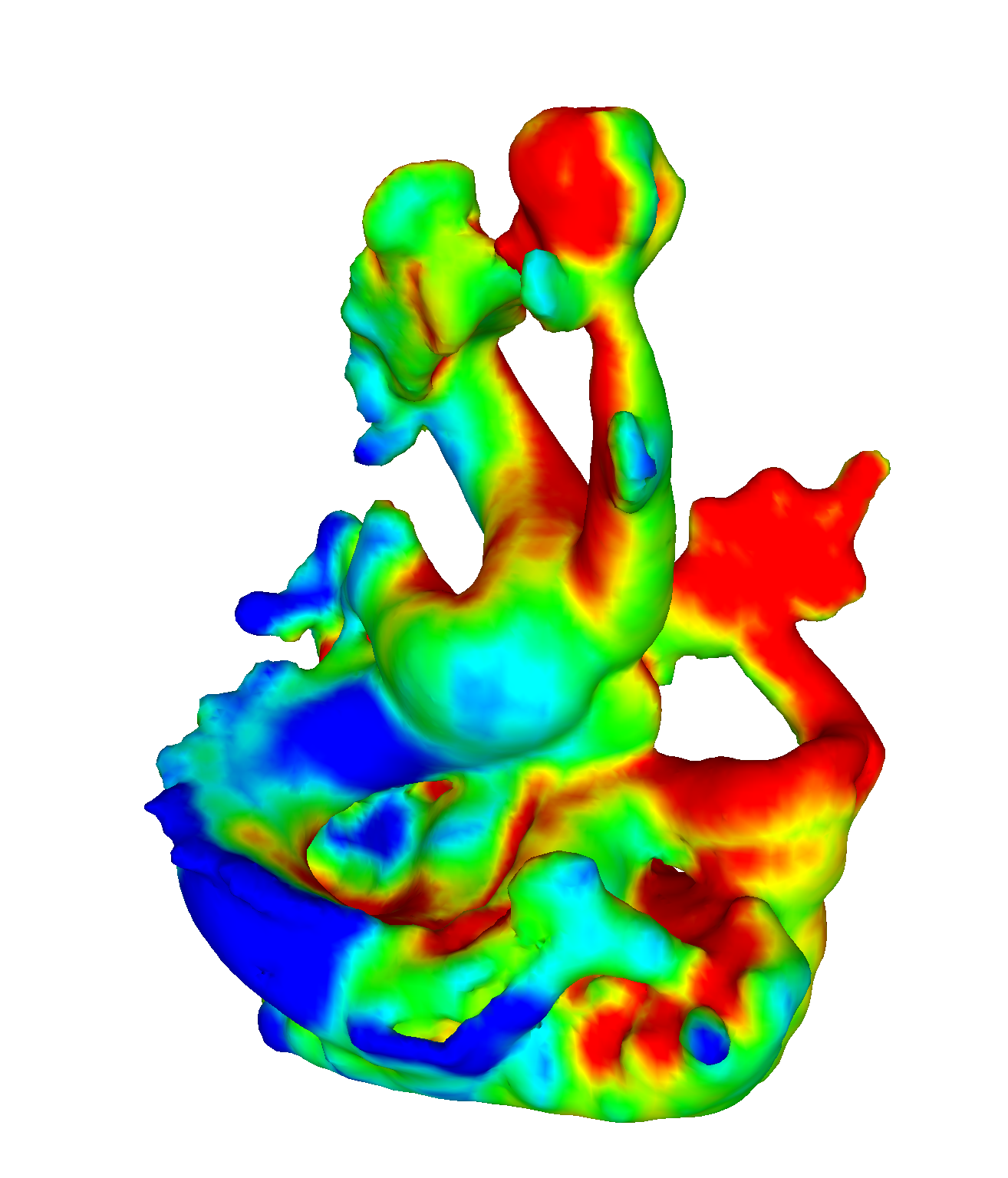}
    \end{minipage}
     &
\begin{minipage}{0.2\textwidth}
      \includegraphics[width=40mm]{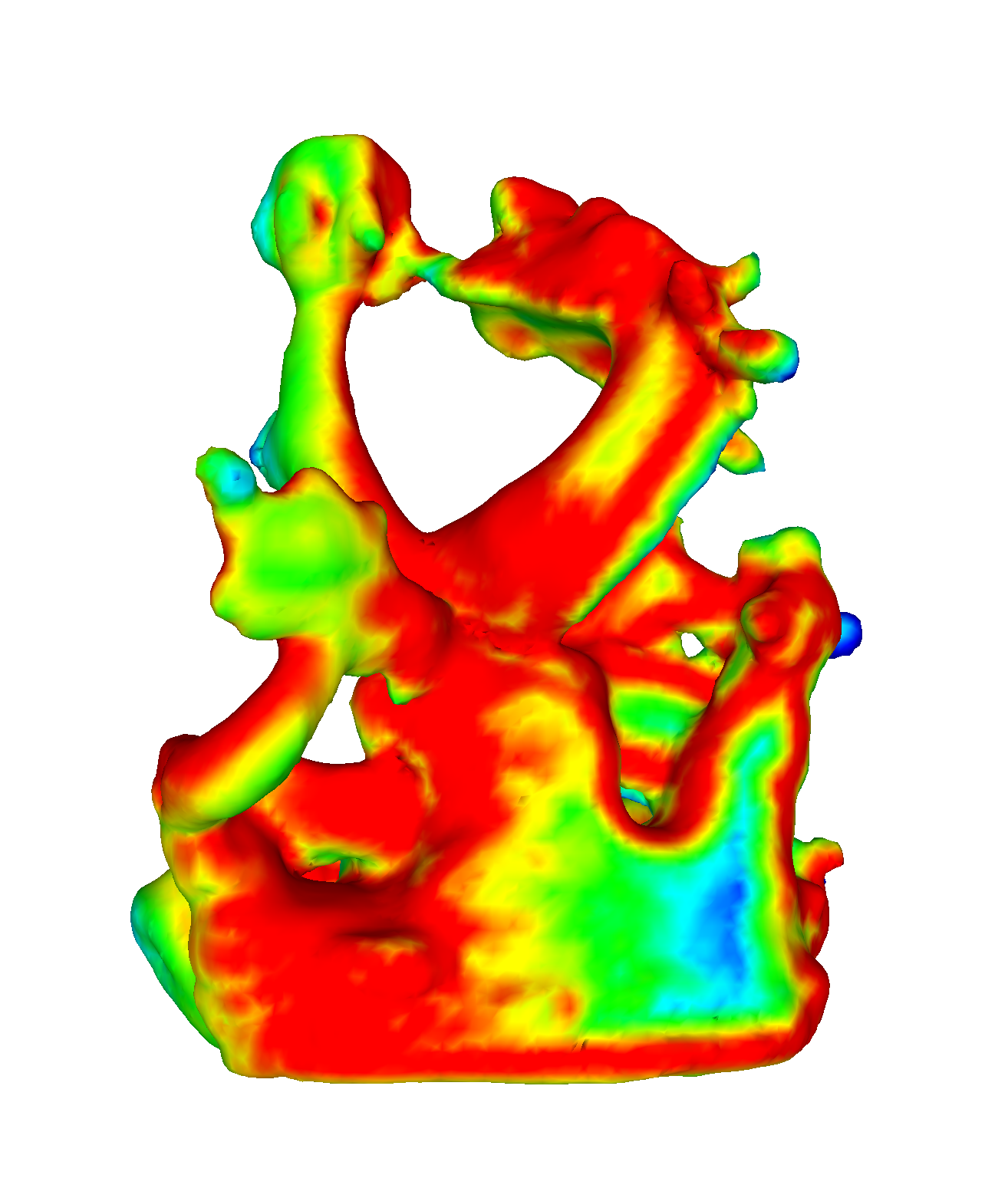}
    \end{minipage}

       \\ \hline

  \end{tabular}}
  \caption{Multi-view reconstruction results comparing to DSS~\cite{Yifan:DSS:2019} and SMVS~\cite{langguth-2016-smvs, Semerjian2014} with the corresponding error visualizations based on Hausdorff distance (red means zero and blue high error).}\label{tbl:dragon}
\end{figure}

\begin{table}[t]
  \centering
  \resizebox{\columnwidth}{!}{
 \begin{tabular}{ | c | c | c | c |}
    \hline
    Object & {\bf Ours} & DSS~\cite{Yifan:DSS:2019} &  SMVS~\cite{langguth-2016-smvs, Semerjian2014}
   \\ \hline
   Torus &  {\bf 0.015637} & 0.035398 &  N/A
   \\ \hline
   Bunny &  {\bf 0.026654} & 0.109432 &  N/A
   \\ \hline
   Dragon & {\bf 0.074723} & 0.179456 & 0.097816
   \\ \hline
  \end{tabular}}
  \caption{Comparison of the symmetric Hausdorff distance between ground truth and reconstructed meshes for torus, bunny and dragon. SMVS could not reconstruct torus and bunny because camera pose estimation failed.}\label{tbl:SHD}
\end{table}

\subsection{Experimental Results}

%\paragraph{Qualitative Results.}

{\bf Qualitative Results.} We compare our results with DSS~\cite{Yifan:DSS:2019}, which is a differentiable renderer for point clouds based on surface splatting~\cite{Zwicker:2001:SS}. We let both systems deform a sphere to fit the target object given as input. When running DSS, we adopt the same settings used in their original experiments: the system randomly selects 12 from a set of 26 views of the target in each optimization cycle, and optimizes for up to 16 cycles. We experimented with different numbers of 3D points and report the best result. For SDFDiff we use our optimization technique from Section~\ref{sec:optimization} using the same set of 26 views. Figure~\ref{tbl:dragon} shows the comparison between SDFDiff and DSS. DSS cannot recover geometric details as accurately as SDFDiff.
%, in particular in the complex Chinese dragon sculpture. 

We also compare our result with SMVS~\cite{langguth-2016-smvs, Semerjian2014}, which is a state-of-the-art shading-aware multi-view 3D reconstruction approach. 
% SMVS works better for complicated geometry, because that gives rich features for the camera pose estimation process. For a better comparison, 
We use the default settings, and provide 1000 randomly sampled views of the dragon rendered by our SDF renderer as input. Note that SMVS automatically recovers camera parameters from the input images and estimates surface albedo and illumination, hence the comparison is not entirely fair. As shown in Figure~\ref{tbl:dragon}, however, even with a large number of input views the SMVS output can be overly smooth and lack details. SMVS may also fail with fewer views due to inaccurate camera pose estimation. In contrast, SDFDiff can obtain better results using only 26 views (with known camera poses, albedo, and illumination). 

%\paragraph{Quantitative Results.}

{\bf Quantitative Results.} Table~\ref{tbl:SHD} compares the symmetric Hausdorff distance between ground truth and reconstructed meshes for torus, bunny and dragon. The visual results of torus and bunny are in supplementary materials. For a fair comparison, we report errors relative to the size of the bounding boxes. We observe that SDFDiff leads to smaller symmetric Hausdorff distances, which means our reconstruction results are closer to the ground truth than the other two approaches.

\begin{figure}[t]
  \centering
  \resizebox{\columnwidth}{!}{
 \begin{tabular}{ | c | c | c | c | c | c | c | c | c | c | c | c | c | c | c | c |}
    \hline 
    init\_res=8 & init\_res=16 & init\_res=32 & init\_res=48 
    \\
    \begin{minipage}{0.098\textwidth}
      \includegraphics[width=17mm]{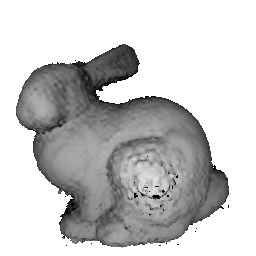}
    \end{minipage}
    &
    \begin{minipage}{0.098\textwidth}
      \includegraphics[width=17mm]{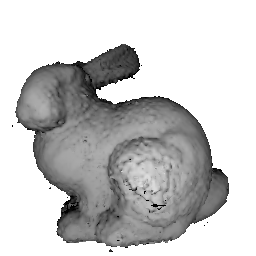}
    \end{minipage}
    &
    \begin{minipage}{0.098\textwidth}
      \includegraphics[width=17mm]{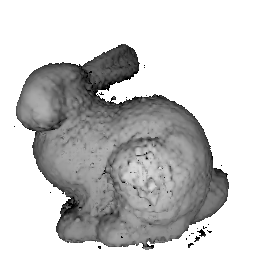}
    \end{minipage}
    &
    \begin{minipage}{0.098\textwidth}
      \includegraphics[width=17mm]{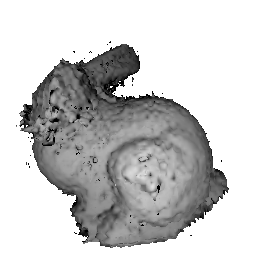}
    \end{minipage}
    % &
    % \begin{minipage}{0.098\textwidth}
    %   \includegraphics[width=17mm]{figures/c1case1_1.png}
    % \end{minipage}
    % &
    % \begin{minipage}{0.098\textwidth}
    %   \includegraphics[width=17mm]{figures/c1case1_2.png}
    % \end{minipage}
    % &
    % \begin{minipage}{0.098\textwidth}
    %   \includegraphics[width=17mm]{figures/c1case1_3.png}
    % \end{minipage}
    \\
    \begin{minipage}{0.098\textwidth}
      \includegraphics[width=17mm]{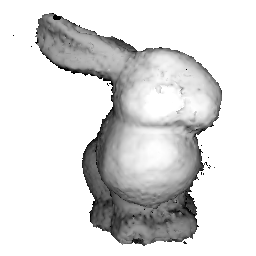}
    \end{minipage}
    &
    \begin{minipage}{0.098\textwidth}
      \includegraphics[width=17mm]{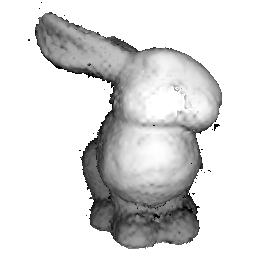}
    \end{minipage}
     &
    \begin{minipage}{0.098\textwidth}
      \includegraphics[width=17mm]{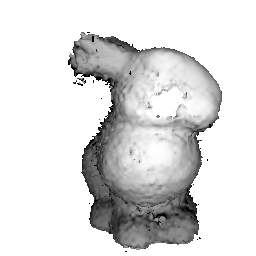}
    \end{minipage}
    &
    \begin{minipage}{0.098\textwidth}
      \includegraphics[width=17mm]{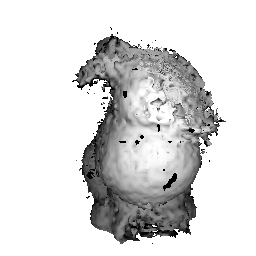}
    \end{minipage}

    \\
    \begin{minipage}{0.098\textwidth}
      \includegraphics[width=17mm]{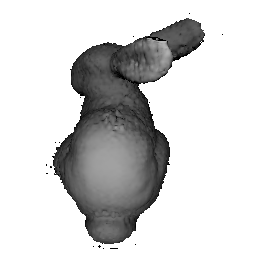}
    \end{minipage}
     &
    \begin{minipage}{0.098\textwidth}
      \includegraphics[width=17mm]{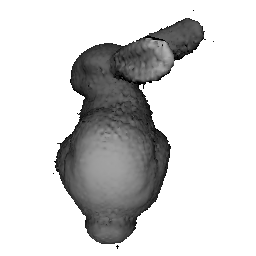}
    \end{minipage}
     &
    \begin{minipage}{0.098\textwidth}
      \includegraphics[width=17mm]{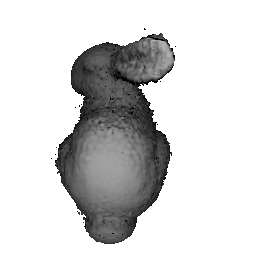}
    \end{minipage}
     &
    \begin{minipage}{0.098\textwidth}
      \includegraphics[width=17mm]{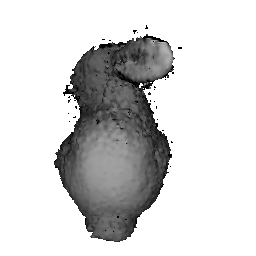}
    \end{minipage}
      \\ \hline
      
    %   init\_res=16 
    % % &
    % % \begin{minipage}{0.098\textwidth}
    % %   \includegraphics[width=17mm]{figures/c1case2_1.png}
    % % \end{minipage}
    % % &
    % % \begin{minipage}{0.098\textwidth}
    % %   \includegraphics[width=17mm]{figures/c1case2_2.png}
    % % \end{minipage}
    % % &
    % % \begin{minipage}{0.098\textwidth}
    % %   \includegraphics[width=17mm]{figures/c1case2_3.png}
    % % \end{minipage}

    %   \\ \hline
      
    %   init\_res=32 
    % % &
    % % \begin{minipage}{0.098\textwidth}
    % %   \includegraphics[width=17mm]{figures/c1case3_1.png}
    % % \end{minipage}
    % % &
    % % \begin{minipage}{0.098\textwidth}
    % %   \includegraphics[width=17mm]{figures/c1case3_2.png}
    % % \end{minipage}
    % % &
    % % \begin{minipage}{0.098\textwidth}
    % %   \includegraphics[width=17mm]{figures/c1case3_3.png}
    % % \end{minipage}

    %   \\ \hline
      
    %   init\_res=48 
    % % &
    % \begin{minipage}{0.098\textwidth}
    %   \includegraphics[width=17mm]{figures/c1case4_1.png}
    % \end{minipage}
    % &
    % \begin{minipage}{0.098\textwidth}
    %   \includegraphics[width=17mm]{figures/c1case4_2.png}
    % \end{minipage}
    % &
    % \begin{minipage}{0.098\textwidth}
    %   \includegraphics[width=17mm]{figures/c1case4_3.png}
    % \end{minipage}

    %   \\ \hline
  \end{tabular}}
  \caption{Given different initial resolutions, with 4 resolution stages, we can find that our 3D reconstruction results are better if the initial resolution is lower.}\label{tbl:c1}
\end{figure}

\begin{figure}[t]
  \centering
  \resizebox{\columnwidth}{!}{
 \begin{tabular}{ | c | c | c | c | c | c | c | c | c | c | c | c | c | c | c | c |}
    \hline 
    1 step & 4 steps & 8 steps & 15 steps
    \\
    % \begin{minipage}{0.1\textwidth}
    % \vspace{0.05mm}
      \includegraphics[width=17mm]{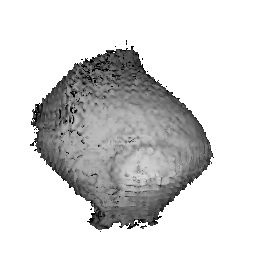}
    % \end{minipage}
    &
    % \begin{minipage}{0.1\textwidth}
    % \vspace{0.05mm}
      \includegraphics[width=17mm]{figures/c1case1_0.png}
    % \end{minipage}
     &
    % \begin{minipage}{0.1\textwidth}
    % \vspace{0.05mm}
      \includegraphics[width=17mm]{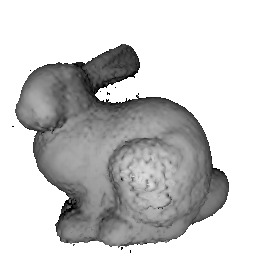}
    %   \vspace{0.05mm}
    % \end{minipage}
       &
    % \begin{minipage}{0.1\textwidth}
    % \vspace{0.05mm}
      \includegraphics[width=17mm]{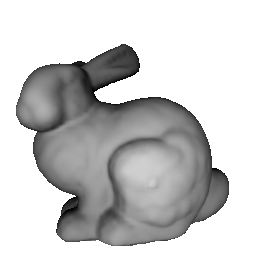}

    \\
    % \begin{minipage}{0.1\textwidth}
    % \vspace{0.05mm}
      \includegraphics[width=17mm]{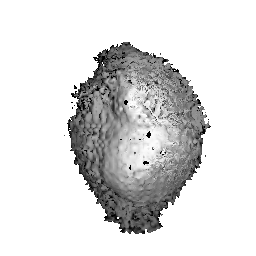}
    % \end{minipage}
     &
    % \begin{minipage}{0.1\textwidth}
    % \vspace{0.05mm}
      \includegraphics[width=17mm]{figures/c1case1_5.png}
    % \end{minipage}
     &
    % \begin{minipage}{0.1\textwidth}
    % \vspace{0.05mm}
      \includegraphics[width=17mm]{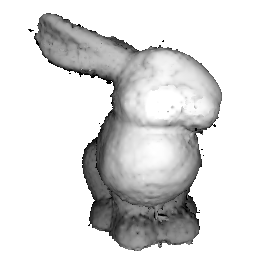}
    %   \vspace{0.05mm}
    % \end{minipage}
     &
    % \begin{minipage}{0.1\textwidth}
    % \vspace{0.05mm}
      \includegraphics[width=17mm]{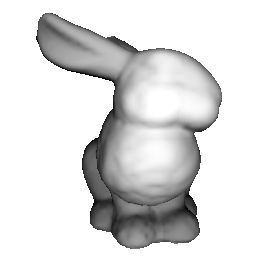}
    % \vspace{0.01mm}
    % \end{minipage}
     
    \\
    % \begin{minipage}{0.1\textwidth}
    % \vspace{0.05mm}
      \includegraphics[width=17mm]{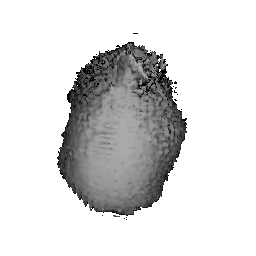}
    % \end{minipage}
     &
    % \begin{minipage}{0.1\textwidth}
    % \vspace{0.05mm}
      \includegraphics[width=17mm]{figures/c1case1_7.png}
    % \end{minipage}
     &
    % \begin{minipage}{0.1\textwidth}
    % \vspace{0.05mm}
      \includegraphics[width=17mm]{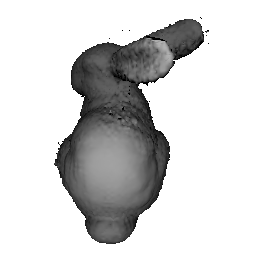}
    %   \vspace{0.05mm}
    % \end{minipage}
     &
    % \begin{minipage}{0.1\textwidth}
    % \vspace{0.05mm}
      \includegraphics[width=17mm]{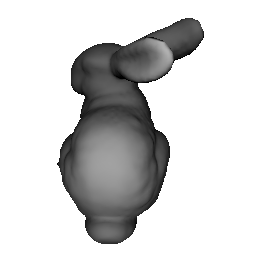}
    %   \vspace{1mm}
    % \vspace{0.01mm}
    % \end{minipage}
      \\ \hline

  \end{tabular}
  }
  \caption{We fix the initial and target resolution to 8 and 64 respectively, but use different numbers of intermediate resolution stages. We find that more resolution stages can give us better results.}\label{tbl:c2}
\end{figure}

\subsection{Parameter Study}

%We conduct several experiments to validate the importance of different parameter settings in our approach.

%\paragraph{Initial Resolution.}
{\bf Initial Resolution.} Figure~\ref{tbl:c1} shows the impact of the initial resolution in our multi-resolution scheme. We fix the number of multi-resolution steps and our target resolution being 64, and then set the initial resolution to be 8, 16, 32, and 48 respectively. We find that a lower initial resolution can reconstruct qualitatively better 3D shapes because it more robustly captures large scale structures.

%\paragraph{Number of Multi-Resolution Steps.}
{\bf Number of Multi-Resolution Steps.} Figure~\ref{tbl:c2} shows that given fixed initial (init\_res=8) and target (target\_res=64) resolutions, adding more multi-resolution steps can give us better results. In particular, single-resolution optimization (1 step) cannot reconstruct the object successfully, further justifying our multi-resolution setup.

\begin{figure}[t]
  \centering
  \resizebox{\columnwidth}{!}{
 \begin{tabular}{ | c | c | c | c | c | c | c | c | c | c | c | c | c | c | c | c |}
    \hline 
    64x64 & 128x128 & 256x256 & 512x512
    
      \\
    % \hline
      
    % \begin{minipage}{0.085\textwidth}
    % \vspace{0.05mm}
      \includegraphics[width=17mm]{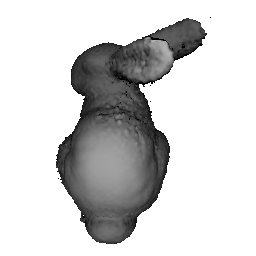}
    % \end{minipage}
    &
    % \vspace{0.05mm}
    % \begin{minipage}{0.085\textwidth}
      \includegraphics[width=17mm]{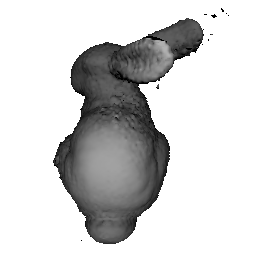}
    % \end{minipage}
    &
    % \begin{minipage}{0.085\textwidth}
    % \vspace{0.05mm}
      \includegraphics[width=17mm]{figures/c2case3_7.png}
    % \end{minipage}
    &
    % \begin{minipage}{0.085\textwidth}
    % \vspace{0.05mm}
      \includegraphics[width=17mm]{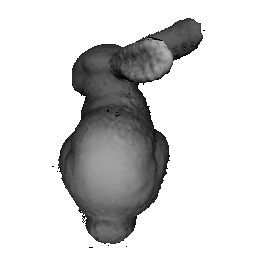}
    % \end{minipage}

      \\ \hline
    
  \end{tabular}
  }
  \caption{We show that the quality of reconstruction results are not affected much by the image resolution.}\label{tbl:c4}
\end{figure}

%\paragraph{Image Resolution.} 
{\bf Image Resolution.} Figure~\ref{tbl:c4} shows that image resolution does not significantly affect the quality of the results, where we use images with various resolutions for optimization.

\begin{figure}[t]
  \centering
  \resizebox{\columnwidth}{!}{
  \begin{tabular}{ | c | c | c | c | c | c | c | c | c | c | c | c | c | c | c | c |}
    \hline 

       \makecell{Noisy \\ views} & Results &  \makecell{Perturbed \\ camera poses} &  Results \\
    % \begin{minipage}{0.12\textwidth}
      \includegraphics[width=17mm]{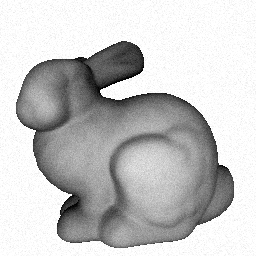}
    % \end{minipage}
    &
    % \begin{minipage}{0.12\textwidth}
      \includegraphics[width=17mm]{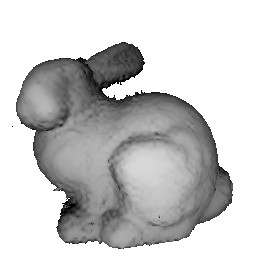}
    % \end{minipage}
    &
    % \begin{minipage}{0.12\textwidth}
      \includegraphics[width=17mm]{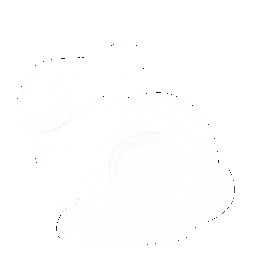}
    % \end{minipage}
    &
    % \begin{minipage}{0.12\textwidth}
      \includegraphics[width=17mm]{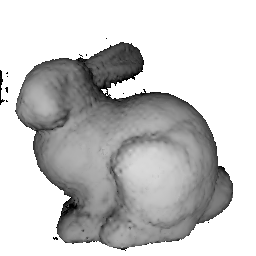}
    % \end{minipage}
    % &
    % \begin{minipage}{0.09\textwidth}
    %   \includegraphics[width=17mm]{figures/26final_cam_64_1.png}
    % \end{minipage}
    % &
    % \begin{minipage}{0.09\textwidth}
    %   \includegraphics[width=17mm]{figures/26final_cam_64_2.png}
    % \end{minipage}
    % &
    % \begin{minipage}{0.09\textwidth}
    %   \includegraphics[width=17mm]{figures/26final_cam_64_3.png}
    % \end{minipage}
    \\
    % \begin{minipage}{0.12\textwidth}
      \includegraphics[width=17mm]{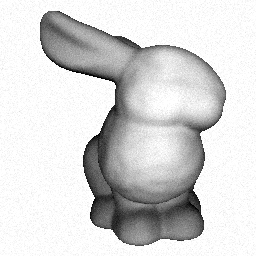}
    % \end{minipage}
     &
    % \begin{minipage}{0.12\textwidth}
      \includegraphics[width=17mm]{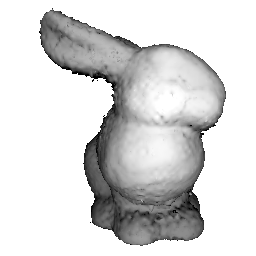}
    % \end{minipage}
     &
    % \begin{minipage}{0.12\textwidth}
      \includegraphics[width=17mm]{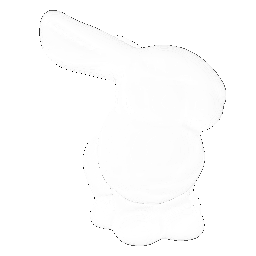}
    % \end{minipage}
    &
    % \begin{minipage}{0.12\textwidth}
      \includegraphics[width=17mm]{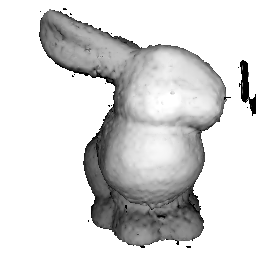}
    % \end{minipage}

    \\
    % \begin{minipage}{0.12\textwidth}
      \includegraphics[width=17mm]{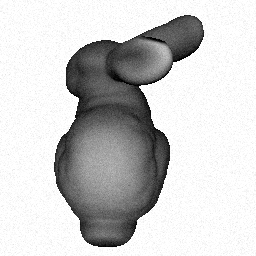}
    % \end{minipage}
     &
    % \begin{minipage}{0.12\textwidth}
      \includegraphics[width=17mm]{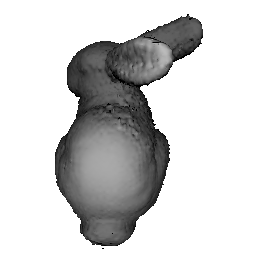}
    % \end{minipage}
    &
    % \begin{minipage}{0.12\textwidth}
      \includegraphics[width=17mm]{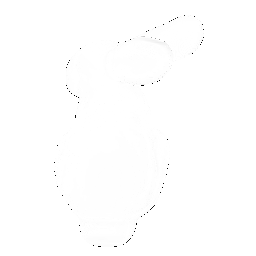}
    % \end{minipage}
      &
    % \begin{minipage}{0.12\textwidth}
      \includegraphics[width=17mm]{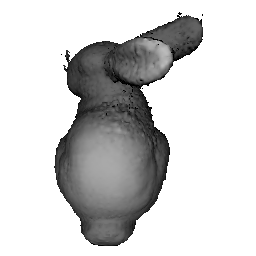}

      \\ \hline
  \end{tabular}}
  \caption{Experimental results with noisy data. All 26 input views or camera poses are perturbed with Gaussian noise (variance=0.03 for views and variance=0.01 for camera poses). The third column shows the view differences caused by perturbed camera poses.}\label{tbl:noisy}
\end{figure}

%\red{\paragraph{{Noisy Data.}} 
{\bf Noisy Data.} As shown in Figure~\ref{tbl:noisy}, when some noise is added to target images or camera poses, our approach can still maintain its robustness.

\begin{table*}[t]
  \centering
  \resizebox{\linewidth}{!}{
 \begin{tabular}{ | c | c | c | c | c | c | c | c | c | c | c | c | c | c | c | c |}
    \hline
     Category &  Airplane & Bench & Cabinet &  Car & Chair & Display & Lamp & Speaker & Rifle & Sofa & Table & Phone & Vessel & Mean \\ \hline
     NMR \cite{kato2018} &  0.6172 &0.4998 &0.7143& 0.7095& 0.4990& 0.5831& 0.4126& 0.6536& 0.6322& 0.6735& 0.4829& 0.7777& 0.5645& 0.6015\\ \hline
    SoftRas (sil.)  \cite{liu2019softras}  &  0.6419 &0.5080 &0.7116 &0.7697& 0.5270& 0.6156& 0.4628&  0.6654 & {\bf 0.6811} &0.6878 &0.4487 &0.7895 &0.5953 &0.6234  \\ \hline
    SoftRas (full) \cite{liu2019softras}&0.6670 &0.5429 &0.7382& 0.7876 &0.5470 &0.6298 &0.4580  &0.6807 & 0.6702& 0.7220 &0.5325 &0.8127 &0.6145 &0.6464\\ \hline
    % DIB-R \cite{chen2019learning} &  0.570  &  0.498  &  0.763  & 0.788  &  0.527  &  0.588 &  0.403& {\bf 0.726}  &  0.561  &  0.677  & 0.508  &  0.743  &  0.609 &  0.612   \\  \hline
    {\bf Ours}  & {\bf 0.6874} & {\bf 0.6860} & {\bf 0.7735} & {\bf 0.8002} & {\bf 0.6436} & {\bf 0.6584} & {\bf 0.5150}& 0.6534 & 0.5553 & {\bf 0.7654} & {\bf 0.6288} & {\bf 0.8278} & {\bf 0.6244} & {\bf 0.6674}  \\ \hline
  \end{tabular}}
  \caption{Comparison of IoU with the state-of-the-art approaches~\cite{kato2018, liu2019softras} on 13 ShapeNet categories.}\label{tbl:iou}
\end{table*}

\begin{figure*}[t]
\includegraphics[width=\textwidth]{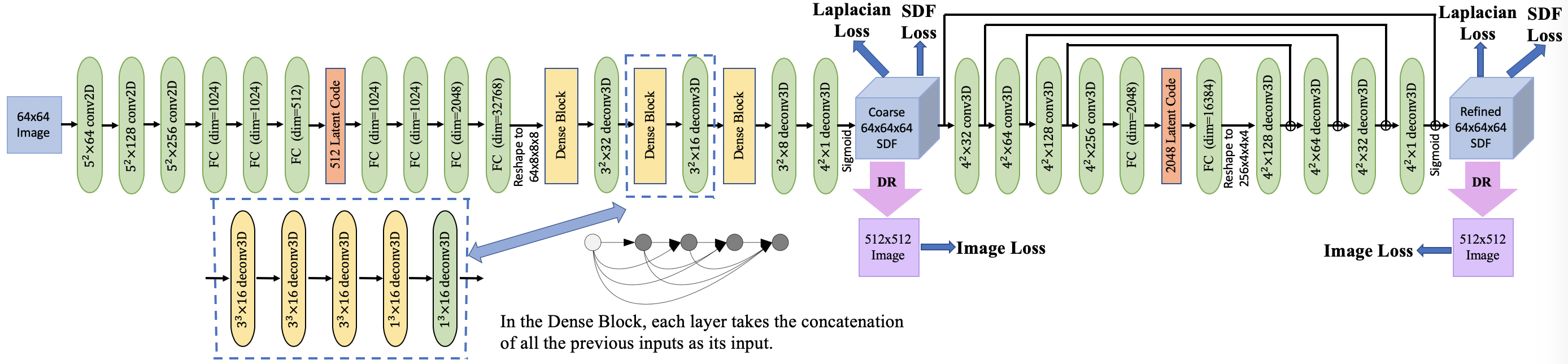}
\caption{Network structure for single-view SDF reconstruction.}
\label{fig:network}
\end{figure*}

\begin{figure*}[t]
  \centering
  \resizebox{\linewidth}{!}{
 \begin{tabular}{ | c | c | c | c | c | c | c | c | c | c | c | c | c | c | c | c |}
    \hline
    % Layout Pattern &   Time Comparison \\ \hline
    
    Method & Input Image & \multicolumn{4}{|c|}{Rendered Views} & Input Image & \multicolumn{4}{|c|}{Rendered Views} \\ \hline
    GT & 
    \begin{minipage}{0.09\textwidth}
      \includegraphics[width=17mm]{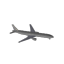}
    \end{minipage}
    &
    \begin{minipage}{0.09\textwidth}
      \includegraphics[width=17mm]{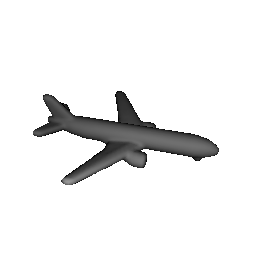}
    \end{minipage}
    &
\begin{minipage}{0.09\textwidth}
      \includegraphics[width=17mm]{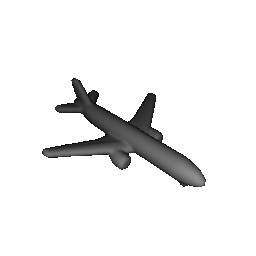}
    \end{minipage}
     &
\begin{minipage}{0.09\textwidth}
      \includegraphics[width=17mm]{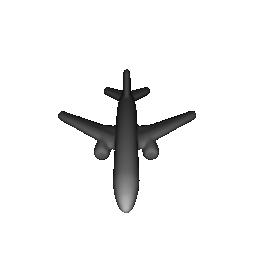}
    \end{minipage}
     &
\begin{minipage}{0.09\textwidth}
      \includegraphics[width=17mm]{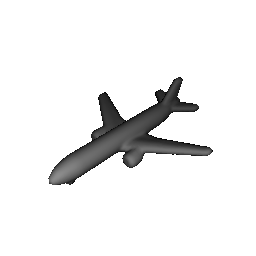}
    \end{minipage}
    & \begin{minipage}{0.09\textwidth}
      \includegraphics[width=17mm]{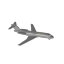}
    \end{minipage}
    &
    \begin{minipage}{0.09\textwidth}
      \includegraphics[width=17mm]{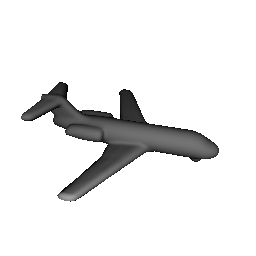}
    \end{minipage}
    &
\begin{minipage}{0.09\textwidth}
      \includegraphics[width=17mm]{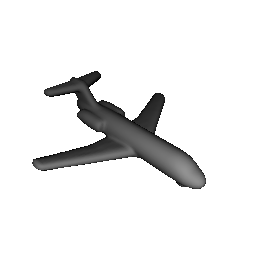}
    \end{minipage}
     &
\begin{minipage}{0.09\textwidth}
      \includegraphics[width=17mm]{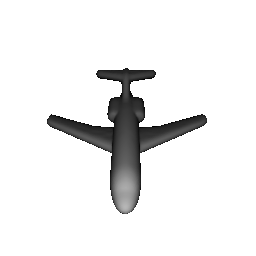}
    \end{minipage}
     &
\begin{minipage}{0.09\textwidth}
      \includegraphics[width=17mm]{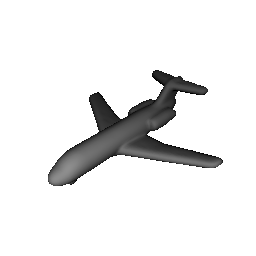}
    \end{minipage}

    \\ \hline
    
    {\bf Ours} & &
    \begin{minipage}{0.09\textwidth}
      \includegraphics[width=17mm]{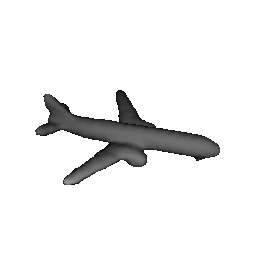}
    \end{minipage}
    &
\begin{minipage}{0.09\textwidth}
      \includegraphics[width=17mm]{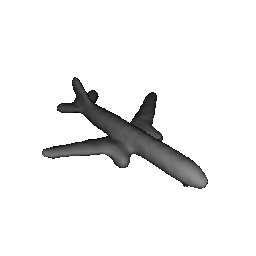}
    \end{minipage}
     &
\begin{minipage}{0.09\textwidth}
      \includegraphics[width=17mm]{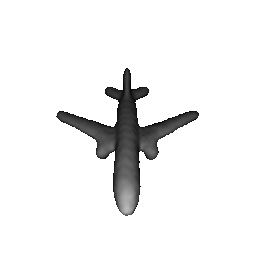}
    \end{minipage}
     &
\begin{minipage}{0.09\textwidth}
      \includegraphics[width=17mm]{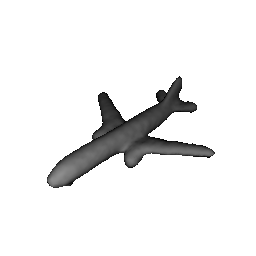}
    \end{minipage}
    & &
    \begin{minipage}{0.09\textwidth}
      \includegraphics[width=17mm]{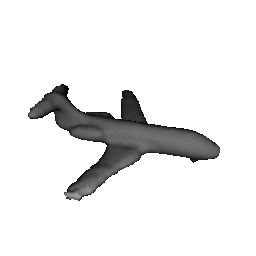}
    \end{minipage}
    &
\begin{minipage}{0.09\textwidth}
      \includegraphics[width=17mm]{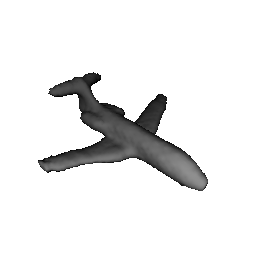}
    \end{minipage}
     &
\begin{minipage}{0.09\textwidth}
      \includegraphics[width=17mm]{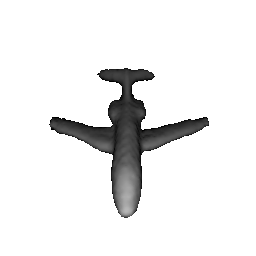}
    \end{minipage}
     &
\begin{minipage}{0.09\textwidth}
      \includegraphics[width=17mm]{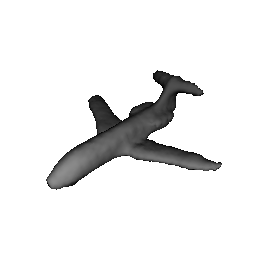}
    \end{minipage}
    
    \\ \hline
    
    SoftRas\cite{liu2019softras} & &
    
     \begin{minipage}{0.09\textwidth}
      \includegraphics[width=17mm]{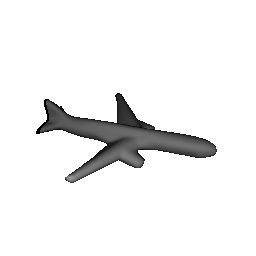}
    \end{minipage}
    &
\begin{minipage}{0.09\textwidth}
      \includegraphics[width=17mm]{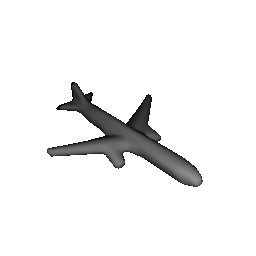}
    \end{minipage}
     &
\begin{minipage}{0.09\textwidth}
      \includegraphics[width=17mm]{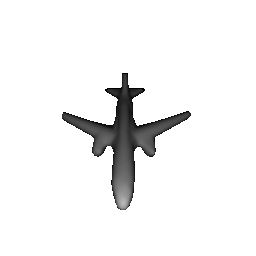}
    \end{minipage}
     &
\begin{minipage}{0.09\textwidth}
      \includegraphics[width=17mm]{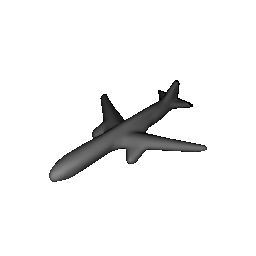}
    \end{minipage}
    & &
    \begin{minipage}{0.09\textwidth}
      \includegraphics[width=17mm]{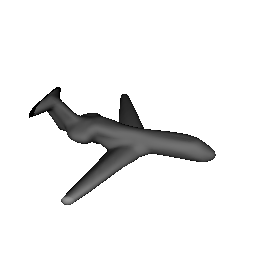}
    \end{minipage}
    &
\begin{minipage}{0.09\textwidth}
      \includegraphics[width=17mm]{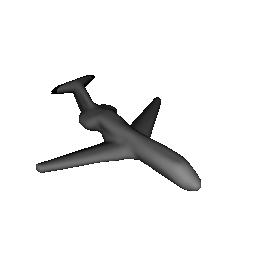}
    \end{minipage}
     &
\begin{minipage}{0.09\textwidth}
      \includegraphics[width=17mm]{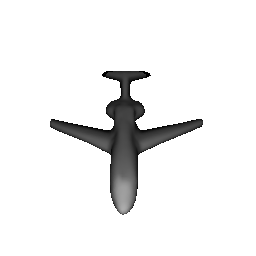}
    \end{minipage}
     &
\begin{minipage}{0.09\textwidth}
      \includegraphics[width=17mm]{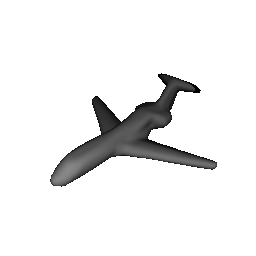}
    \end{minipage}

  \\ \hline

          GT & \begin{minipage}{0.09\textwidth}
      \includegraphics[width=17mm]{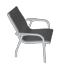}
    \end{minipage}
    &
    \begin{minipage}{0.09\textwidth}
      \includegraphics[width=17mm]{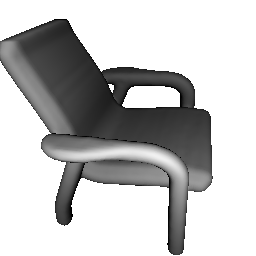}
    \end{minipage}
    &
\begin{minipage}{0.09\textwidth}
      \includegraphics[width=17mm]{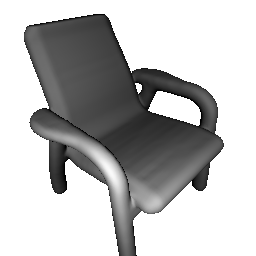}
    \end{minipage}
     &
\begin{minipage}{0.09\textwidth}
      \includegraphics[width=17mm]{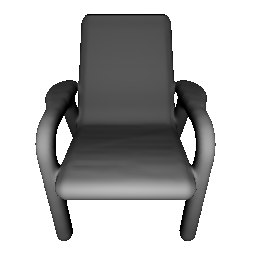}
    \end{minipage}
     &
\begin{minipage}{0.09\textwidth}
      \includegraphics[width=17mm]{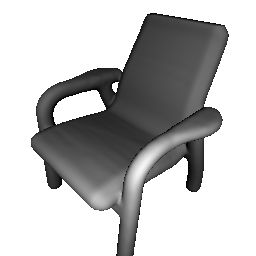}
    \end{minipage}
    & \begin{minipage}{0.09\textwidth}
      \includegraphics[width=17mm]{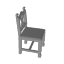}
    \end{minipage}
    &
    \begin{minipage}{0.09\textwidth}
      \includegraphics[width=17mm]{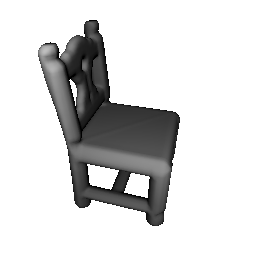}
    \end{minipage}
    &
\begin{minipage}{0.09\textwidth}
      \includegraphics[width=17mm]{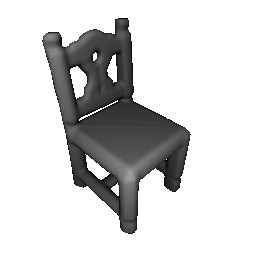}
    \end{minipage}
     &
\begin{minipage}{0.09\textwidth}
      \includegraphics[width=17mm]{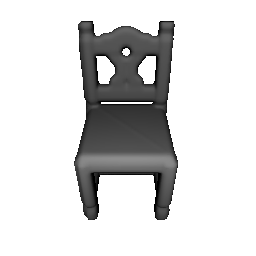}
    \end{minipage}
     &
\begin{minipage}{0.09\textwidth}
      \includegraphics[width=17mm]{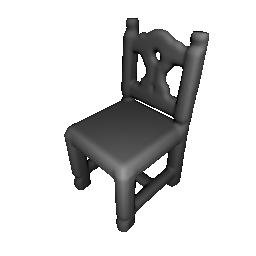}
    \end{minipage}

    \\ \hline
    
    {\bf Ours} & &
    \begin{minipage}{0.09\textwidth}
      \includegraphics[width=17mm]{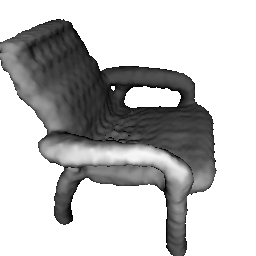}
    \end{minipage}
    &
\begin{minipage}{0.09\textwidth}
      \includegraphics[width=17mm]{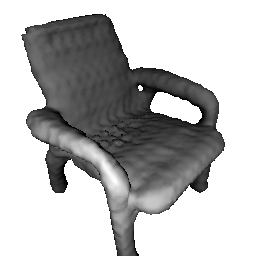}
    \end{minipage}
     &
\begin{minipage}{0.09\textwidth}
      \includegraphics[width=17mm]{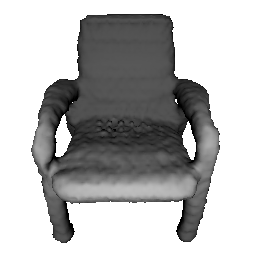}
    \end{minipage}
     &
\begin{minipage}{0.09\textwidth}
      \includegraphics[width=17mm]{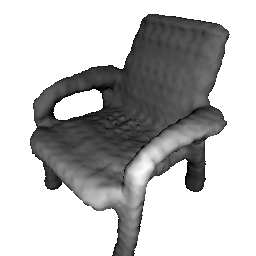}
    \end{minipage}
    & &
    \begin{minipage}{0.09\textwidth}
      \includegraphics[width=17mm]{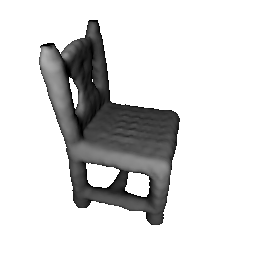}
    \end{minipage}
    &
\begin{minipage}{0.09\textwidth}
      \includegraphics[width=17mm]{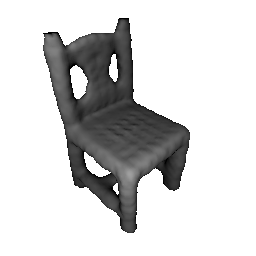}
    \end{minipage}
     &
\begin{minipage}{0.09\textwidth}
      \includegraphics[width=17mm]{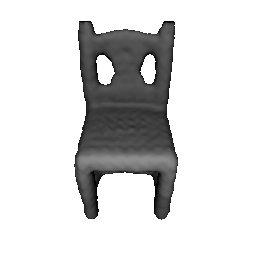}
    \end{minipage}
     &
\begin{minipage}{0.09\textwidth}
      \includegraphics[width=17mm]{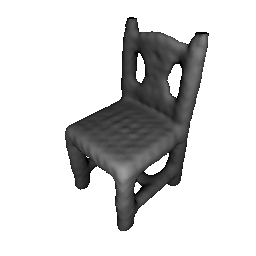}
    \end{minipage}

        \\ \hline
    
    SoftRas\cite{liu2019softras} & &
    
    \begin{minipage}{0.09\textwidth}
      \includegraphics[width=17mm]{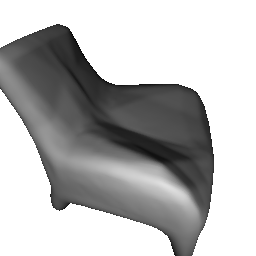}
    \end{minipage}
    &
\begin{minipage}{0.09\textwidth}
      \includegraphics[width=17mm]{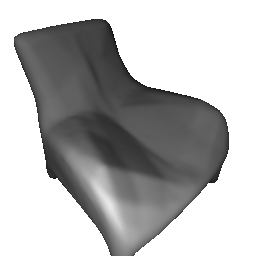}
    \end{minipage}
     &
\begin{minipage}{0.09\textwidth}
      \includegraphics[width=17mm]{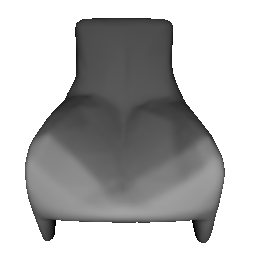}
    \end{minipage}
     &
\begin{minipage}{0.09\textwidth}
      \includegraphics[width=17mm]{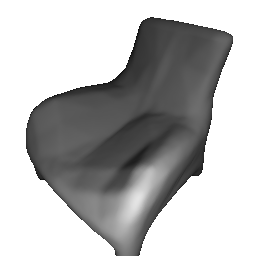}
    \end{minipage}
    & &
    \begin{minipage}{0.09\textwidth}
      \includegraphics[width=17mm]{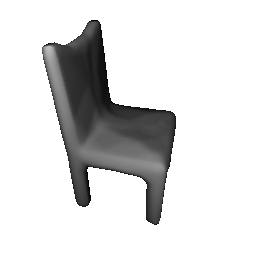}
    \end{minipage}
    &
\begin{minipage}{0.09\textwidth}
      \includegraphics[width=17mm]{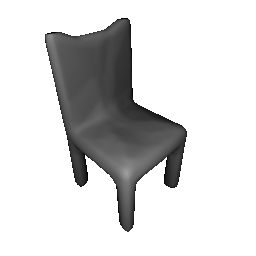}
    \end{minipage}
     &
\begin{minipage}{0.09\textwidth}
      \includegraphics[width=17mm]{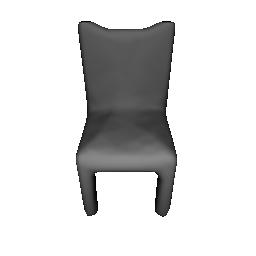}
    \end{minipage}
     &
\begin{minipage}{0.09\textwidth}
      \includegraphics[width=17mm]{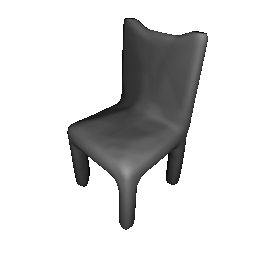}
    \end{minipage}

\\ \hline

     GT & \begin{minipage}{0.09\textwidth}
      \includegraphics[width=17mm]{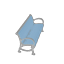}
    \end{minipage}
    &
    \begin{minipage}{0.09\textwidth}
      \includegraphics[width=17mm]{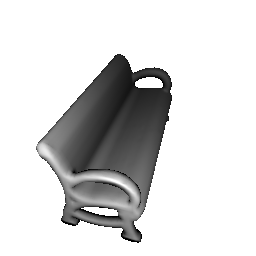}
    \end{minipage}
    &
\begin{minipage}{0.09\textwidth}
      \includegraphics[width=17mm]{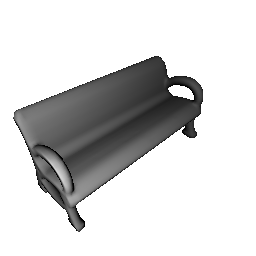}
    \end{minipage}
     &
\begin{minipage}{0.09\textwidth}
      \includegraphics[width=17mm]{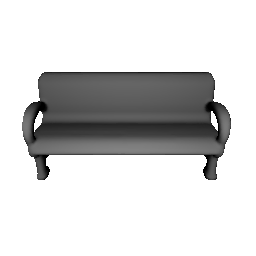}
    \end{minipage}
     &
\begin{minipage}{0.09\textwidth}
      \includegraphics[width=17mm]{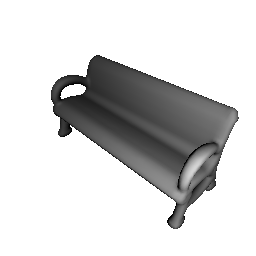}
    \end{minipage}
    & 
    \begin{minipage}{0.09\textwidth}
      \includegraphics[width=17mm]{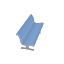}
    \end{minipage}
    &
    \begin{minipage}{0.09\textwidth}
      \includegraphics[width=17mm]{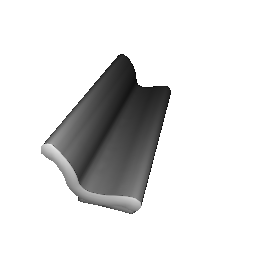}
    \end{minipage}
    &
\begin{minipage}{0.09\textwidth}
      \includegraphics[width=17mm]{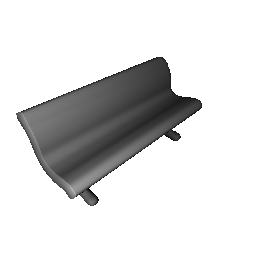}
    \end{minipage}
     &
\begin{minipage}{0.09\textwidth}
      \includegraphics[width=17mm]{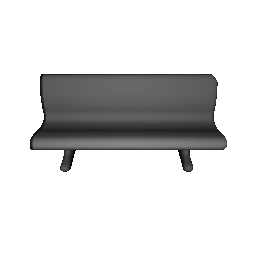}
    \end{minipage}
     &
\begin{minipage}{0.09\textwidth}
      \includegraphics[width=17mm]{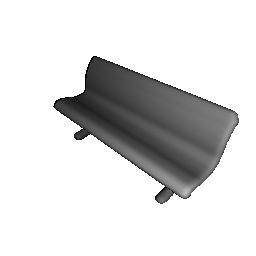}
    \end{minipage}

    \\ \hline
    
    {\bf Ours} & &
    \begin{minipage}{0.09\textwidth}
      \includegraphics[width=17mm]{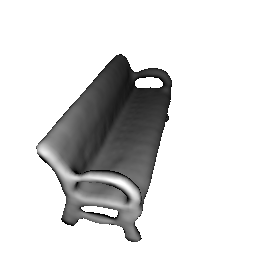}
    \end{minipage}
    &
\begin{minipage}{0.09\textwidth}
      \includegraphics[width=17mm]{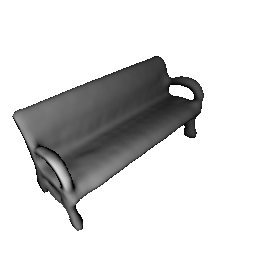}
    \end{minipage}
     &
\begin{minipage}{0.09\textwidth}
      \includegraphics[width=17mm]{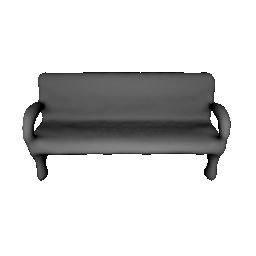}
    \end{minipage}
     &
\begin{minipage}{0.09\textwidth}
      \includegraphics[width=17mm]{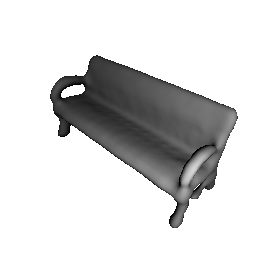}
    \end{minipage}
    & &
    \begin{minipage}{0.09\textwidth}
      \includegraphics[width=17mm]{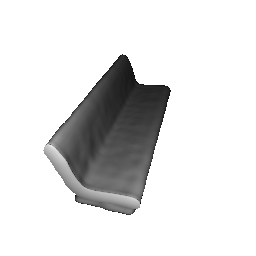}
    \end{minipage}
    &
\begin{minipage}{0.09\textwidth}
      \includegraphics[width=17mm]{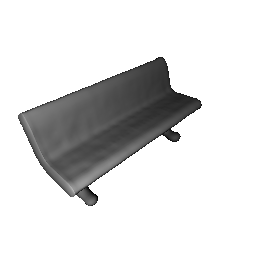}
    \end{minipage}
     &
\begin{minipage}{0.09\textwidth}
      \includegraphics[width=17mm]{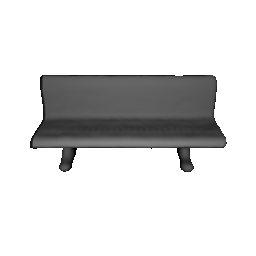}
    \end{minipage}
     &
\begin{minipage}{0.09\textwidth}
      \includegraphics[width=17mm]{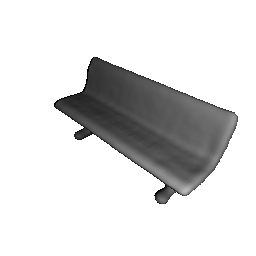}
    \end{minipage}
  
\\ \hline
    
    SoftRas\cite{liu2019softras} & &
    \begin{minipage}{0.09\textwidth}
      \includegraphics[width=17mm]{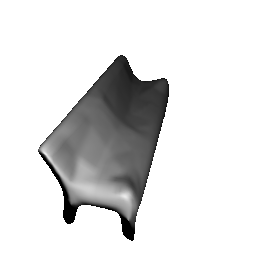}
    \end{minipage}
    &
\begin{minipage}{0.09\textwidth}
      \includegraphics[width=17mm]{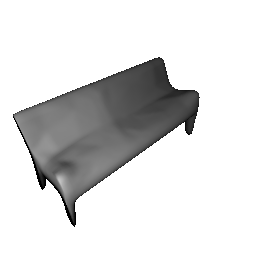}
    \end{minipage}
     &
\begin{minipage}{0.09\textwidth}
      \includegraphics[width=17mm]{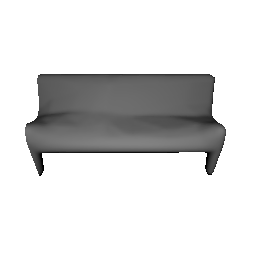}
    \end{minipage}
     &
\begin{minipage}{0.09\textwidth}
      \includegraphics[width=17mm]{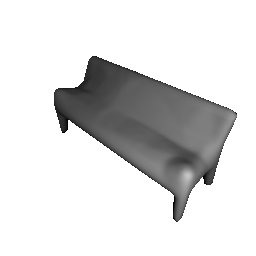}
    \end{minipage}
    & &
    \begin{minipage}{0.09\textwidth}
      \includegraphics[width=17mm]{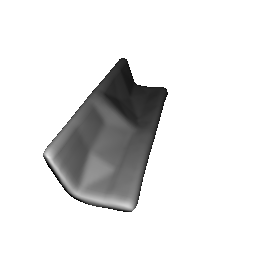}
    \end{minipage}
    &
\begin{minipage}{0.09\textwidth}
      \includegraphics[width=17mm]{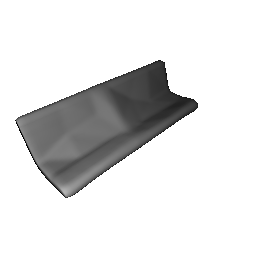}
    \end{minipage}
     &
\begin{minipage}{0.09\textwidth}
      \includegraphics[width=17mm]{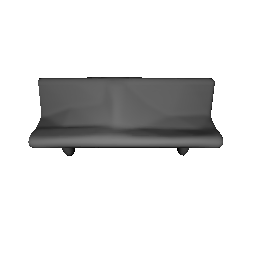}
    \end{minipage}
     &
\begin{minipage}{0.09\textwidth}
      \includegraphics[width=17mm]{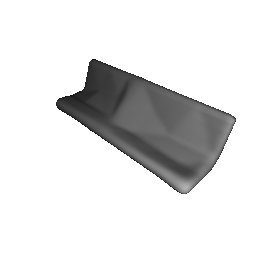}
    \end{minipage}

    \\ \hline

    GT & \begin{minipage}{0.09\textwidth}
      \includegraphics[width=17mm]{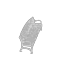}
    \end{minipage}
    &
    \begin{minipage}{0.09\textwidth}
      \includegraphics[width=17mm]{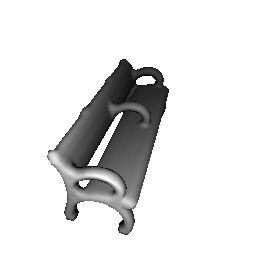}
    \end{minipage}
    &
\begin{minipage}{0.09\textwidth}
      \includegraphics[width=17mm]{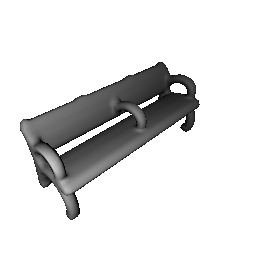}
    \end{minipage}
     &
\begin{minipage}{0.09\textwidth}
      \includegraphics[width=17mm]{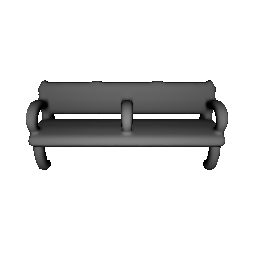}
    \end{minipage}
     &
\begin{minipage}{0.09\textwidth}
      \includegraphics[width=17mm]{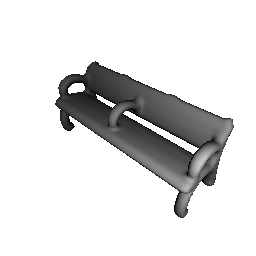}
    \end{minipage}
    & \begin{minipage}{0.09\textwidth}
      \includegraphics[width=17mm]{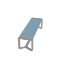}
    \end{minipage}
    &
    \begin{minipage}{0.09\textwidth}
      \includegraphics[width=17mm]{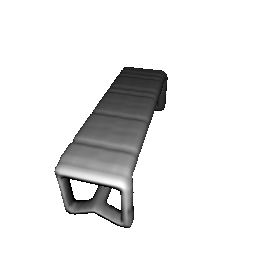}
    \end{minipage}
    &
\begin{minipage}{0.09\textwidth}
      \includegraphics[width=17mm]{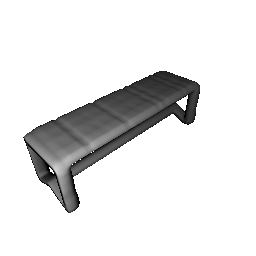}
    \end{minipage}
     &
\begin{minipage}{0.09\textwidth}
      \includegraphics[width=17mm]{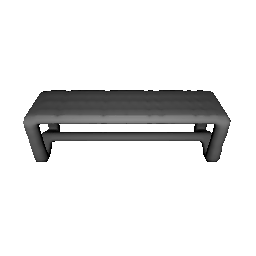}
    \end{minipage}
     &
\begin{minipage}{0.09\textwidth}
      \includegraphics[width=17mm]{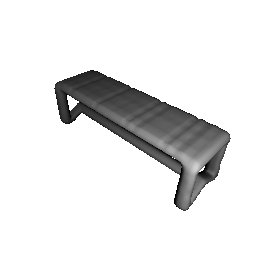}
    \end{minipage}

    \\ \hline
    
    {\bf Ours} & &
    \begin{minipage}{0.09\textwidth}
      \includegraphics[width=17mm]{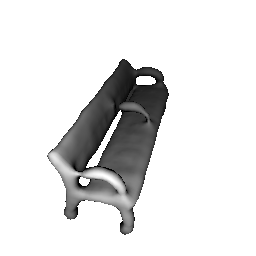}
    \end{minipage}
    &
\begin{minipage}{0.09\textwidth}
      \includegraphics[width=17mm]{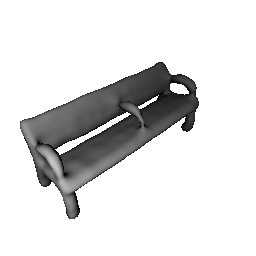}
    \end{minipage}
     &
\begin{minipage}{0.09\textwidth}
      \includegraphics[width=17mm]{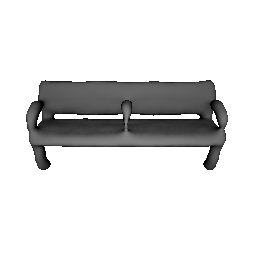}
    \end{minipage}
     &
\begin{minipage}{0.09\textwidth}
      \includegraphics[width=17mm]{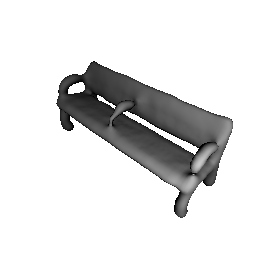}
    \end{minipage}
    & &
    \begin{minipage}{0.09\textwidth}
      \includegraphics[width=17mm]{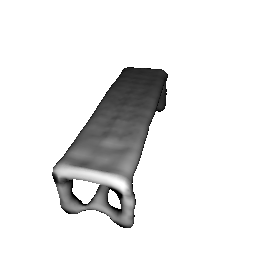}
    \end{minipage}
    &
\begin{minipage}{0.09\textwidth}
      \includegraphics[width=17mm]{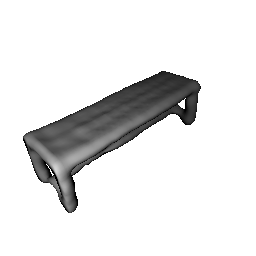}
    \end{minipage}
     &
\begin{minipage}{0.09\textwidth}
      \includegraphics[width=17mm]{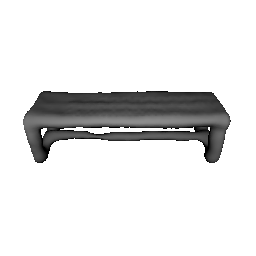}
    \end{minipage}
     &
\begin{minipage}{0.09\textwidth}
      \includegraphics[width=17mm]{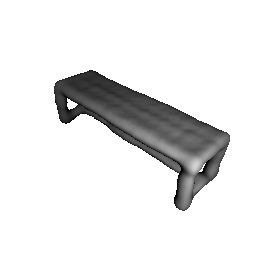}
    \end{minipage}

   \\ \hline
    
    SoftRas\cite{liu2019softras} & &
    \begin{minipage}{0.09\textwidth}
      \includegraphics[width=16.5mm]{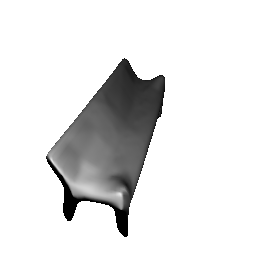}
    \end{minipage}
    &
\begin{minipage}{0.09\textwidth}
      \includegraphics[width=16.5mm]{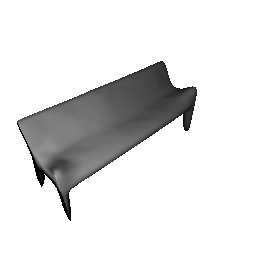}
    \end{minipage}
     &
\begin{minipage}{0.09\textwidth}
      \includegraphics[width=16.5mm]{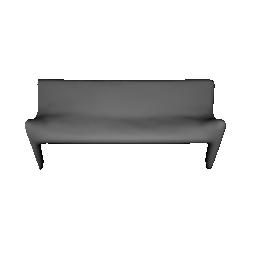}
    \end{minipage}
     &
\begin{minipage}{0.09\textwidth}
      \includegraphics[width=16.5mm]{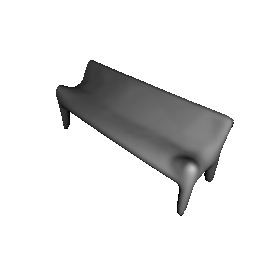}
    \end{minipage}
    & &
    \begin{minipage}{0.09\textwidth}
      \includegraphics[width=16.5mm]{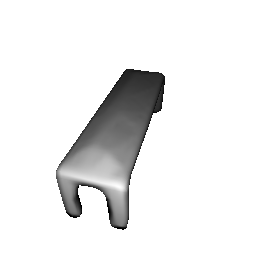}
    \end{minipage}
    &
\begin{minipage}{0.09\textwidth}
      \includegraphics[width=16.5mm]{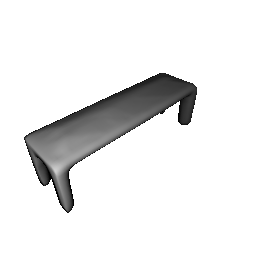}
    \end{minipage}
     &
\begin{minipage}{0.09\textwidth}
      \includegraphics[width=16.5mm]{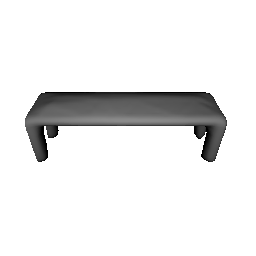}
    \end{minipage}
     &
\begin{minipage}{0.09\textwidth}
      \includegraphics[width=16.5mm]{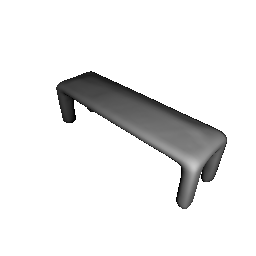}
    \end{minipage}

    \\ \hline

     GT & \begin{minipage}{0.09\textwidth}
      \includegraphics[width=16.5mm]{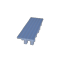}
    \end{minipage}
    &
    \begin{minipage}{0.09\textwidth}
      \includegraphics[width=16.5mm]{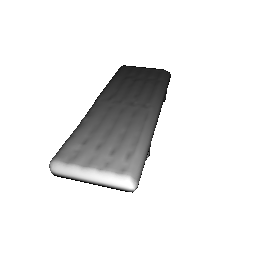}
    \end{minipage}
    &
\begin{minipage}{0.09\textwidth}
      \includegraphics[width=16.5mm]{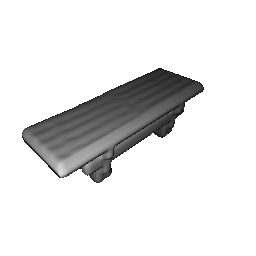}
    \end{minipage}
     &
\begin{minipage}{0.09\textwidth}
      \includegraphics[width=16.5mm]{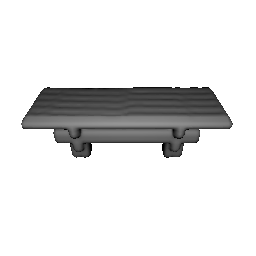}
    \end{minipage}
     &
\begin{minipage}{0.09\textwidth}
      \includegraphics[width=16.5mm]{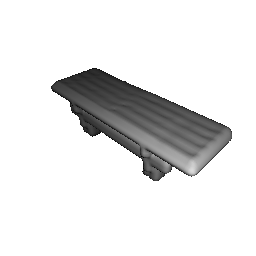}
    \end{minipage}
    & \begin{minipage}{0.09\textwidth}
      \includegraphics[width=16.5mm]{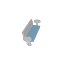}
    \end{minipage}
    &
    \begin{minipage}{0.09\textwidth}
      \includegraphics[width=16.5mm]{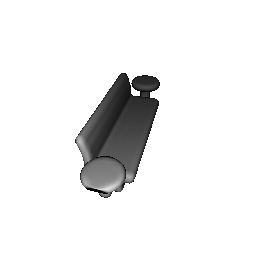}
    \end{minipage}
    &
\begin{minipage}{0.09\textwidth}
      \includegraphics[width=16.5mm]{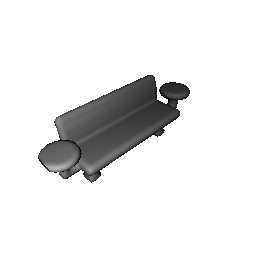}
    \end{minipage}
     &
\begin{minipage}{0.09\textwidth}
      \includegraphics[width=16.5mm]{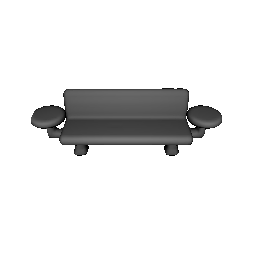}
    \end{minipage}
     &
\begin{minipage}{0.09\textwidth}
      \includegraphics[width=16.5mm]{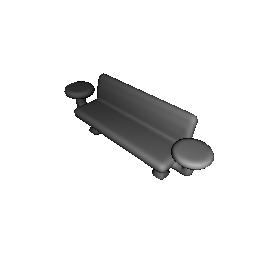}
    \end{minipage}

    \\ \hline
    
    {\bf Ours} & &
    \begin{minipage}{0.085\textwidth}
      \includegraphics[width=16.5mm]{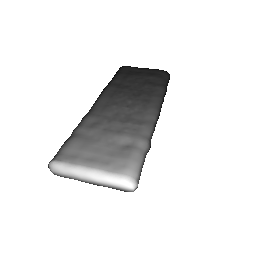}
    \end{minipage}
    &
\begin{minipage}{0.085\textwidth}
      \includegraphics[width=16.5mm]{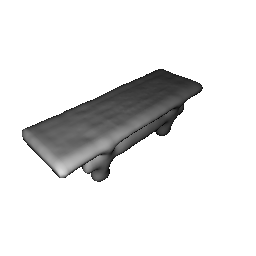}
    \end{minipage}
     &
\begin{minipage}{0.085\textwidth}
      \includegraphics[width=16.5mm]{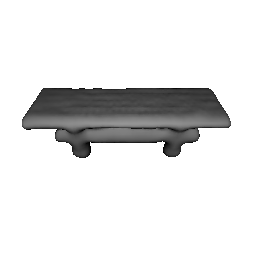}
    \end{minipage}
     &
\begin{minipage}{0.085\textwidth}
      \includegraphics[width=16.5mm]{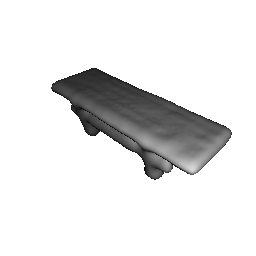}
    \end{minipage}
    & &
    \begin{minipage}{0.085\textwidth}
      \includegraphics[width=16.5mm]{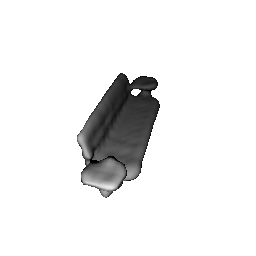}
    \end{minipage}
    &
\begin{minipage}{0.085\textwidth}
      \includegraphics[width=16.5mm]{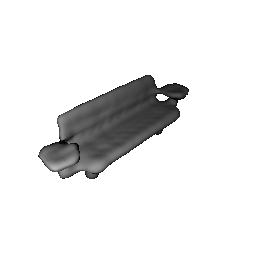}
    \end{minipage}
     &
\begin{minipage}{0.085\textwidth}
      \includegraphics[width=16.5mm]{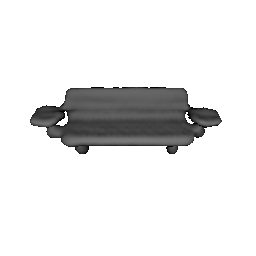}
    \end{minipage}
     &
\begin{minipage}{0.085\textwidth}
      \includegraphics[width=16.5mm]{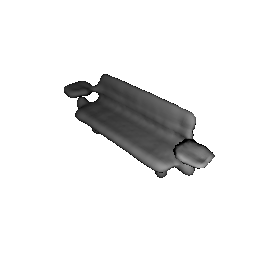}
    \end{minipage}

   \\ \hline
    
    SoftRas\cite{liu2019softras} & &
    \begin{minipage}{0.085\textwidth}
      \includegraphics[width=16.5mm]{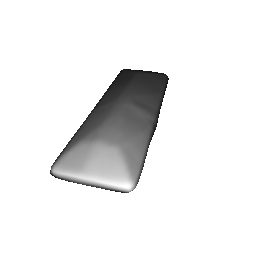}
    \end{minipage}
    &
\begin{minipage}{0.085\textwidth}
      \includegraphics[width=16.5mm]{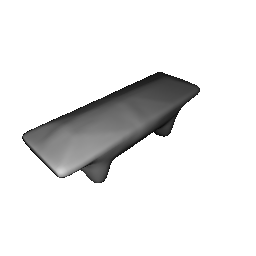}
    \end{minipage}
     &
\begin{minipage}{0.085\textwidth}
      \includegraphics[width=16.5mm]{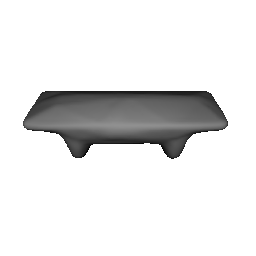}
    \end{minipage}
     &
\begin{minipage}{0.085\textwidth}
      \includegraphics[width=16.5mm]{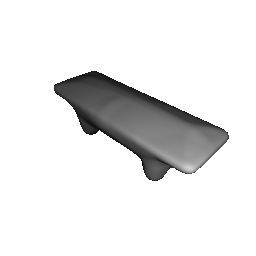}
    \end{minipage}
    & &
    \begin{minipage}{0.085\textwidth}
      \includegraphics[width=16.5mm]{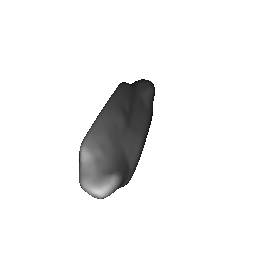}
    \end{minipage}
    &
\begin{minipage}{0.085\textwidth}
      \includegraphics[width=16.5mm]{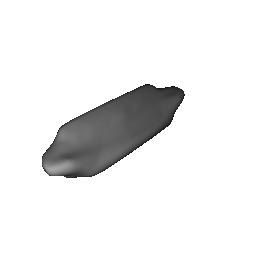}
    \end{minipage}
     &
\begin{minipage}{0.085\textwidth}
      \includegraphics[width=16.5mm]{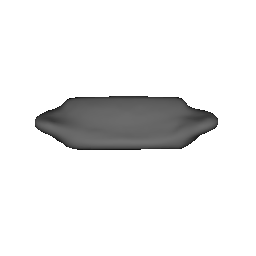}
    \end{minipage}
     &
\begin{minipage}{0.085\textwidth}
      \includegraphics[width=16.5mm]{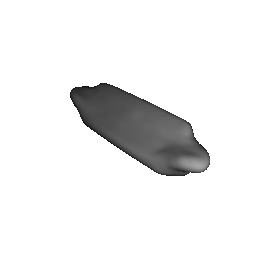}
    \end{minipage}

\\ \hline

  \end{tabular}}
  \caption{Single-view reconstruction results for airplanes, chairs, and benches.}\label{tbl:compare1}
\end{figure*}

\section{Learning-based Single-view Reconstruction}

%\subsection{Dataset and Experimental Setup}

In the following experiments, we leverage SDFDiff to train a neural network to perform single-view 3D reconstruction without 3D supervision. We use the same dataset as \cite{kato2018, liu2019softras}, which includes 13 categories of objects from ShapeNet \cite{Chang2015ShapeNet}. Each object has 24 rendered images from different views at $64\times 64$ resolution. We use the same train/validate/test sets on the same dataset as in \cite{kato2018, liu2019softras, yan2016perspective}, and the standard reconstruction metric, {\em i.e.,} 3D intersection over union (IoU)~\cite{liu2019softras} for quantitative comparisons.

%\paragraph{Network.}

{\bf Network.} Our network contains two parts as shown in Figure~\ref{fig:network}. The first part is an Encoder-Decoder network which takes images as input and outputs coarse SDF results. The second part is a refiner network to further improve the quality of the 3D reconstruction results. The network is trained on all the 3D shapes in the dataset simultaneously.

%\paragraph{Loss Function.}

{\bf Loss Function.} In addition to the energy function as shown in Section~\ref{sec:energyfunction} containing the $L_2$ image-based loss $\mathcal{L}_{\mathrm{img}}$ and the SDF loss $\mathcal{L}_{\mathrm{reg}}$ ensuring the SDF values represent a valid signed distance function, we also add a geometry loss $\mathcal{L}_{\mathrm{geo}}$ that regularizes the finite difference Laplacian of the predicted SDFs to obtain smooth outputs. Furthermore, we use a narrow band technique to control the effects of the SDF and Laplacian losses since we care more about these losses locally around the surfaces. Also, the SDF-loss cannot be enforced everywhere on the discrete grid due to singularities (e.g., the medial axis of the shape forms a sharp crease) in the continuous SDF. The narrow-band considers the SDF-loss only close to the surface, avoiding SDF discretization issues elsewhere in the volume. We use a distance-based binary mask $\mathcal{M}$ to zero them out further away from the zero level-set. The mask is defined as
\begin{equation}
% \label{eq:invrendering}
\mathcal{M} = ||SDF|| \leq \mu \times voxelSize,
\end{equation}
where $\mu$ is a hyperparameter to define the width of the narrow band. We currently set it to be 1.6, which is determined experimentally.
The final loss function is a weighted sum of the three losses with weights $\lambda_1=\lambda_2=0.02$,
\begin{equation}
% \label{eq:invrendering}
\mathcal{L} = \mathcal{L}_{\mathrm{img}} + \mathcal{M} \otimes (\lambda_1 \mathcal{L}_{\mathrm{reg}} + \lambda_2 \mathcal{L}_{\mathrm{geo}}).
\end{equation}

%\paragraph{Training Process.}

{\bf Training Process.} We first train the Encoder-Decoder part of the network alone based on the three loss terms. Then we fix the encoder and decoder and train the refiner network on the same three loss terms to get refined SDF shapes. In the end, we train all the three parts, {\em i.e., encoder, decoder, and refiner} together to further improve the results. We do not use the multi-resolution approach.

%\paragraph{Qualitative Evaluation.}

{\bf Qualitative Evaluation.} \autoref{tbl:compare1} %and \autoref{tbl:compare2} 
shows that our method can reconstruct detailed objects and accurately recover complicated topologies. In contrast, SoftRasterizer~\cite{liu2019softras} relies on a template mesh with spherical topology and it cannot capture the complex topology of the chairs.

%\paragraph{Quantitative Evaluation.}

{\bf Quantitative Evaluation.} We compare our method with the state-of-the-art~\cite{kato2018, liu2019softras} in terms of 3D IoU scores in Table~\ref{tbl:iou}. Our method can reconstruct shapes with finer details in the 13 categories. In addition, the IoU numbers show that our results achieve higher accuracy, where our scores surpass other approaches in most of the categories. A comparison to Chen et al.~\cite{NIPS2019_9156} is omitted because they use different data preprocessing than the other methods~\cite{kato2018, liu2019softras}.

\section{Discussion and Limitations}

As a main advantage, SDFs can represent arbitrary topologies, in contrast to triangle meshes that are restricted to the topology of a template. In contrast to point clouds, SDFs inherently represent continuous watertight surfaces. We demonstrated applications of our approach in multi-view shape reconstruction, and single view 3D reconstruction using deep learning. Our experimental results showed that we can more robustly perform multi-view reconstruction than a state-of-the-art point-based differentiable renderer. In addition, we achieve state-of-the-art results on single view 3D reconstruction with deep learning models. 

In our multi-view 3D reconstruction approach, our current shading model is not sufficient to perform inverse rendering from real images taken with a camera. For example, we currently do not include effects such as shadows, interreflections, texture, non-diffuse surfaces, or complex illumination. In contrast to rasterization-based differentiable renderers, our ray tracing-based renderer could be extended to include all such effects. A disadvantage of our deep learning approach is that we output a discrete SDF on a 3D grid. Instead, we could learn a continuous signed distance function represented by a deep network like in DeepSDF~\cite{Park2019}. This would be more memory efficient, but it might be computationally too expensive for unsupervised 3D reconstruction with differentiable rendering, since it would require to evaluate the network for each ray marching step. 

\section{Conclusion}

We proposed a novel approach to differentiable rendering using signed distance functions to represent watertight 3D geometry. Our rendering algorithm is based on sphere tracing, but we observe that only the local shading computation needs to be differentiable in our framework, which makes the approach computationally more efficient and allows for straightforward integration into deep learning frameworks. We demonstrate applications in multi-view 3D reconstruction and unsupervised single-view 3D reconstruction using deep neural networks. Our experimental results illustrate the advantages over geometry representations such as point clouds and meshes. In particular, we report the state-of-the-art results in shape reconstruction.

\section{Acknowledgements}
This work builds on initial explorations of multi-view 3D reconstruction with differentiable rendering~\cite{simone} by Simone Raimondi and the last author. This project was supported by NSF IIS grant nr. $\#1813583$. We also appreciate active discussion with Qian Zheng, Hui Huang, and Daniel Cohen-Or at Shenzhen University.
{\small
\bibliographystyle{ieee_fullname}
\bibliography{egpaper_final}
}

\end{document}